# Efficient Structure from Motion for Oblique UAV Images Based on Maximal Spanning Tree Expansions

**San Jiang** [1], **Wanshou Jiang** [1,*]

[1] State Key Laboratory of Information Engineering in Surveying, Mapping and Remote Sensing, Wuhan University, Wuhan 430072, China; jiangsan870211@whu.edu.cn (S.J.)
* Correspondence: jws@whu.edu.cn; Tel.: +86-27-6877-8092 (ext. 8321)

**Abstract:** The primary contribution of this paper is an efficient Structure from Motion (SfM) solution for oblique unmanned aerial vehicle (UAV) images. First, an algorithm, considering spatial relationship constrains between image footprints, is designed for match pair selection with assistant of UAV flight control data and oblique camera mounting angles. Second, a topological connection network (TCN), represented by an undirected weighted graph, is constructed from initial match pairs, which encodes overlap area and intersection angle into edge weights. Then, an algorithm, termed MST-Expansion, is proposed to extract the match graph from the TCN where the TCN is firstly simplified by a maximum spanning tree (MST). By further analysis of local structure in the MST, expansion operations are performed on the nodes of the MST for match graph enhancement, which is achieved by introducing critical connections in two expansion directions. Finally, guided by the match graph, an efficient SfM solution is proposed, and its validation is verified through comprehensive analysis and comparison using three UAV datasets captured with different oblique multi-camera systems. Experiment results demonstrate that the efficiency of image matching is improved with a speedup ratio ranging from 19 to 35, and competitive orientation accuracy is achieved from both relative bundle adjustment (BA) without GCPs (Ground Control Points) and absolute BA with GCPs. At the same time, images in the three datasets are successfully oriented. For orientation of oblique UAV images, the proposed method can be a more efficient solution.

## 1. Introduction

Oblique images are becoming increasingly more important in the photogrammetry community. Compared with the traditional photogrammetry, oblique imaging technology is capable of capturing both footprints and facades of targets, which enables and simplifies identification and interpretation of some hard-to-see facilities from the unique perspective view (Remondino and Gerke, 2015). This technique can be considered as the bridge between classical and terrestrial acquisitions (Sun et al., 2016), and their abilities for civil applications have been more and more documented (Gerke, 2009; Vetrivel et al., 2015; Zhu et al., 2015). For some situations, e.g. power line inspection (Jiang et al., 2017), prerequisites for flexible data acquisition and high resolution images are required and airplane-based platforms could not satisfy the needs. Nowadays, unmanned aerial vehicle (UAV) has emerged as a new data acquisition platform (Colomina and Molina, 2014), which features rapid data acquisition, low cost and easy to use. Because of relative lower flight altitude, UAV-based photogrammetric systems could effectively capture and provide images with extremely high spatial resolutions and have been used in fields of agricultural management (Aicardi et al., 2016b), engineering monitoring (Matsuoka et al., 2012) and topological survey (Gonçalves and Henriques, 2015), etc. Therefore, the combination of UAV platforms and oblique photogrammetric systems is rational for strength integration, which improves both of their applicability for acquisitions (Aicardi et al., 2016a; Lin et al., 2015).

Although advantages come from the UAV oblique imaging systems, new challenges are posed to post-processing of oblique UAV images. First, very precise geo-referencing devices for direct image orientation, including GNSS (Global Navigation Satellite System) receivers and IMU (Inertial Measurement Unit) sensors, could not be used onboard mainly because of the payload limitation of UAV platforms and high costs of the devices. Even with lightweight and low-cost sensors integrated, the orientation data is generally regarded as approximation of camera poses (Nex and Remondino, 2014). Second, compared with the traditional vertical photogrammetry, a larger number of images with higher overlap degree in both along-track and across-track directions are captured from UAV oblique imaging systems because of the adoption of extra oblique cameras (Rupnik et al., 2013). In addition, the spatial resolution of the captured images is extremely high because of the lower flight altitude. Consequently, efficient image orientation is mandatorily required to resume accurate camera poses prior to further processing, including dense matching and 3D reconstruction.

To resume camera poses, the traditional photogrammetric methods cannot satisfy the demand of oblique UAV images due to the fact that good initial values for camera poses and 3D points are need to be known prior to scene reconstruction (Remondino and Gerke, 2015). However, the situation frequently occurs where an UAV system can just provide rough pose data from an automatic cruising system, which is not precise enough to be used as a priori camera poses. Fortunately, a cutting-edge technique, termed 'Structure from Motion' (SfM), has emerged, which differs fundamentally from the conventional photogrammetry (Westoby et al., 2012). In contrast, SfM requires neither camera poses nor 3D points to be known prior to scene reconstruction, and it can automatically and simultaneously solves all these unknown parameters from overlapped images (Snavely et al., 2006). The strengths of this technique has been verified and reported in comparison tests with terrestrial laser scanning (Zhang et al., 2016) and in applications of architectural model reconstruction (Ippoliti et al., 2015), landslide monitoring (Chidburee et al., 2016), etc. These merits provide advantages for orientation of oblique UAV images. However, the primal SfM is originally designed in the field of computer vision for the purpose of recovering small scale scenes or scenarios recorded by small size photos with low resolution. Thus, high computational costs are required for oblique UAV images due to the large data size and the very high spatial resolution.

In the literature, many approaches have been proposed to improve the efficiency of SfM, especially for image matching (Rupnik et al., 2013) and bundle adjustment (Sun et al., 2016). Among these researches, more attention has been devoted to the acceleration of image matching, which was comprehensively reviewed in (Hartmann et al., 2015). An efficient and robust approach is to search match pairs prior to actual feature matching execution, and existed solutions can be categorized into two groups. For the first group, without any prior knowledge about camera poses, image retrieval using vocabulary tree is the standard method to conduct coarse matching for image similarity ranking. For each image, feature matching is then applied to fewer images with highest similarity scores (Agarwal et al., 2010; Heinly et al., 2015). Although vocabulary tree is the most widely used technique to choose visually similar image pairs, the number of similar items is hard to determine and mismatches are inevitably existed due to symmetric and repetitive patterns. Meanwhile, the time consumption for vocabulary tree build is not satisfying because of the large volume and the high resolution of UAV images. For the second group, the spatial adjacent relationship is the most obvious clue suitable for match pair selection. Some approaches have been documented in recent years. In the work of (AliAkbarpour et al., 2015), a consecutive frame-to-frame matching strategy was adopted for sequential images, where the temporal consistency constrain was used to reduce the combinatory complexity of images and then improve the efficiency of feature matching. With the assistant of coarse GNSS/IMU priors, image footprints can be calculated and utilized for match pair selection based on the criterion that whether or not the footprints of one image pair are overlapped. (Irschara et al., 2011) proposed a POS-assisted solution for match pair selection. First, visually overlapped image pairs were chosen using a modified vocabulary tree strategy; second, the overlap criterion was used to filter out false match pairs. Similarly, for automatic orientation of oblique aerial images, (Rupnik et al., 2013, 2014) employed a new image connectivity graph to speed up image matching. Overlap area and camera looking direction were tested to establish image links, which restricts feature matching for suitable image pairs. Compared with image retrieval based methods, the prior camera poses form a more direct and reliable clue for match pair searching.

However, vast redundant match pairs are still existed in the match pairs generated from the mentioned approaches due to the fact that match pairs are determined merely based on a direct adjacent principle (Xu et al., 2016), where any image pair with overlap region is marked as candidates for feature matching. To achieve further reduction, some researches have focused on the analysis of the entire topology of match pairs. Initial match pairs are firstly utilized to construct a topological connection network (TCN), which is usually represented by an undirected graph. Then, topological analysis is conducted to remove non-essential edges from the graph. For example, a skeletal graph extraction method was proposed to implement an efficient SfM for large-scale and unordered datasets in the research of (Snavely et al., 2008), where a modified maximal leaf spanning tree was used to select a subset of images. By using the bag-of-words method to choose visually overlapped images, (Havlena et al., 2010) created an undirected graph and selected a small set of images by computing an approximate minimal connected dominating set from an initial TCN graph. (Alsadik et al., 2014) found a minimum camera network by iteratively adjusting camera orientation, which minimizes expected orientation errors to a required precision. For match pair selection, most of these methods are designed for photos from internet community and applied for images with small size and low resolution. Considering different characteristics between community photos and UAV images, improvements can be dramatically achieved from the prior GNSS/IMU data. Recently, (Xu et al., 2016) embedded a skeletal camera network (SCN) into a SfM pipeline to enable efficient 3D reconstruction for highly overlapped UAV images. Image footprints were firstly calculated from flight control data. Then, a TCN was constructed with edge weighted by overlap area of image pairs. Finally, a hierarchical degree bounded maximum spanning tree (MST) method was used to generate the SCN within an iterative procedure. With the analysis of the image connection topology, initial match pairs can be dramatically reduced for further improvement of image matching efficiency.

However, some issues can be noticed: (1) the algorithms developed in most of the previous studies do not consider the characteristics of oblique UAV images, since they were initially designed aimed at vertical images. For vertical images, the number of matched features is almost positive proportional to the overlap area for one image pair, but for oblique images feature matching is dramatically influenced by the intersection angle of the image pair due to the fact that most feature extraction algorithms have limited tolerance to the projective deformations caused by oblique imaging angles. (2) Although match pairs are reduced using the topological analysis of TCN, image connections within strips are excessively emphasized because of larger overlap degree and smaller intersection angle. However, it is essential to retain image connections across strips for most photogrammetry missions, in order to confirm a reliable image connection network in BA. (3) Exhausted intersection test is widely required to check whether or not two footprints are overlapped because it is hard to determine a fixed searching radius for neighbor retrieval. With image volume increasing, the time complexity for the pairwise test is quadratic in the number of images. Overall, above mentioned solutions are not suitable to be directly applied onto oblique UAV images.

This paper proposes an efficient strategy for match pair selection and designs a match graph extraction algorithm to guide image matching within a SfM solution. First, a match pair selection strategy, independent on a fixed distance threshold for neighbor retrieval and with linear computational costs, is designed by the further analysis of the spatial relationship constrains. Second, a TCN is constructed from initial match pairs with edges weighted by both overlap area and intersection angle. Then, a non-iterative and two-stage algorithm is proposed for match graph extraction from the TCN. Finally, comprehensive analysis and comparison for the proposed solution are conducted using oblique UAV images. The overall flowchart and experimental strategy in this study is illustrated in Figure 1.

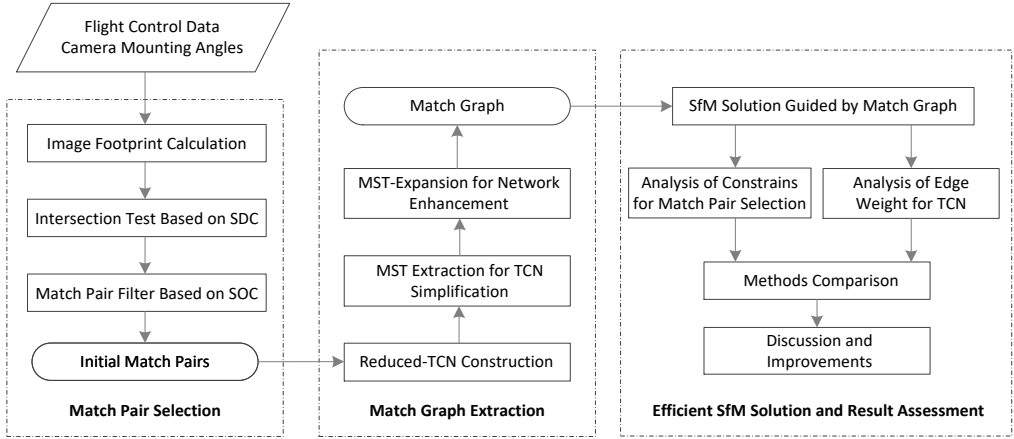

**Figure 1.** The overall flowchart and experimental strategy of the proposed SfM solution.

This paper is organized as follows. Section 2 details the match pair selection strategy. The match graph extraction algorithm is presented in section 3, which is followed by the design of a match graph guided SfM solution in Section 4. Comprehensive analysis and comparison of the efficient SfM are presented in Section 5. In addition, some aspects are discussed in Section 6. Finally, Section 7 presents the conclusion.

**2. Spatial Relationship Constrained Match Pair Selection**

The purpose of match pair selection is to choose overlapped image pairs, which is an efficient and robust strategy to decrease the complexity of pairwise image matching. For UAV photogrammetry missions, the GNSS/IMU data from the onboard flight control system can be used as a priori poses to facilitate match pair selection. Image footprints are firstly computed and intersection tests are executed to determine whether or not the footprints are overlapped. The image pair would be retained if overlapped region is existed; otherwise, be pruned. In the literature, the solution based on exhausted tests (Xu et al., 2016) is preferred to the method using a fixed radius for neighbor searching (Rupnik et al., 2013) because it is hard to choose the distance threshold. However, the time complexity is quadratic in the number of images. In this section, an algorithm, termed SRC-InterTest, is proposed for match pair selection from oblique UAV images, which is described in Algorithm 1. Compared with existed methods, the main improvements are introduced by exploiting two extra spatial constrains. The first constrain is SDC (Spatial Distance Constrain) used to achieve a more efficient intersection test, which is independent on a fixed radius for neighbor searching and can avoid exhausted tests. The second constrain is SOC (Spatial Overlay Constrain), which filters image pairs with too narrow or too small overlapped regions to decrease the risk of incorrect matches involving. The algorithm is designed and implemented as follows.

*2.1. Intersection Test Considering Spatial Distance Constrain*

Camera poses are calculated before intersection tests. For oblique UAV photogrammetry, the onboard GNSS/IMU data actually provides the positions and orientations of the platform and camera poses can be parameterized with the platform poses and the relative poses, which are respective displacements between the platform and oblique cameras (Sun et al., 2016). Suppose that a multi-camera oblique system with $m$ cameras is adopted and the relative poses are estimated and represented by $C_i = \{r_i, t_i\}$, where $r$ is a rotational matrix and $t$ is a translation vector. Similarly, $n$ poses, measuring positions and orientations of the platform at exposure stations, are expressed by $C_j = \{R_j, T_j\}$. Then, the pose for camera $i$ at exposure station $j$ can be calculated by Equation(1).

$$C_i^j = \{r_i R_j, R_j^T t_i + T_j\} \quad (1)$$

where the first and the second items of $C_i^j$ stand for the rotational matrix and the translation vector, respectively; $i$ range from 1 to $m$; $j$ ranges from 1 to $n$. Remarkable, platform positions measured in the geographic reference system are transformed to a rectangular coordinate system and platform orientations recorded in the navigation system are converted to the photogrammetric system.

After calculation of camera poses, three major steps consist of an intersection test. First, image footprints are computed by projecting image corners to an average elevation plane. Second, a spatial index tree, indexing central locations of the footprints, is built using the K-nearest neighbors algorithm (Cover and Hart, 1967), and nearest neighbors for each location are searched and sorted in ascending order. Finally, intersection tests are executed sequentially between the target location and its neighboring locations. Considering SDC, reasonable termination conditions are determined to avoid exhausted intersection tests by using a four-quadrant indicator and a non-intersection indicator.

The structure of these two indicators is illustrated in Figure 2, where the red and blue colored columns construct the four-quadrant and non-intersection indicators, respectively. For the four-quadrant indicator, one element, indexed from column one to four, is set as one when current intersection test in the corresponding quadrant is true; otherwise, be set as zero.

The fifth element is a counter that counts the number of elements valued zero from the first to the fourth columns. For the non-intersection indicator, the first element is a flag, which indicates whether or not non-intersection tests have continuously occurred. The second element is also a counter, which accumulates the count of continuous non-intersection tests.

|   | four-quadrant indicator | | | | | non-inter indicator | |
|---|---|---|---|---|---|---|---|
|   | 0 | 0 | 0 | 0 | 4 | FALSE | 0 |
| 1 | 0 | 1 | 0 | 0 | 3 | FALSE | 0 |
| 2 | 1 | 1 | 0 | 0 | 2 | FALSE | 0 |
| 3 | 1 | 1 | 0 | 1 | 1 | FALSE | 0 |
| 4 | 0 | 1 | 0 | 1 | 2 | TRUE | 1 |
| … | … | … | … | … | … | … | … |
| N | 0 | 0 | 0 | 0 | 4 | TRUE | 4 |
|   | flag | | | | counter | flag | counter |

**Figure 2.** Structure of the four-quadrant indicator and the non-intersection indicator.

For one intersection test, assume that the target and neighboring footprints are $g_t$ and $g_n$, respectively; the quadrant index of $g_n$ with respect to $g_t$ is denoted as $i$, which ranges from 1 to 4. If $g_n$ intersects with $g_t$, then $g_n \cap g_t \neq \varnothing$; otherwise, $g_n \cap g_t = \varnothing$. The four-quadrant indicator $Q$ and the non-intersection indicator $N$ are updated according to Equation(2) and Equation(3), respectively. Meanwhile, the counter $Q_5$ of the four-quadrant indicator would be decreased or increased by one according to the conditions in Equation(2). The counter $N_{count}$ is set as zero when $g_n$ intersects with $g_t$; otherwise, be increased by one. The intersection test would be stopped when $Q_5 = 4$ or $N_{count}$ exceeds a specified threshold $T_I$. In Figure 2, the termination condition reaches in row N with $Q_5 = 4$. Thus, based on the SDC constrain, an intersection test can be rationally interrupted without the need to specify a distance threshold, and exhausted tests can be avoid with the satisfaction that intersection tests are sufficiently conducted to choose all overlapped image pairs.

$$Q_i = \begin{cases} 1, & g_n \cap g_t \neq \varnothing \text{ and } Q_i = 0 \\ 0, & g_n \cap g_t = \varnothing \text{ and } Q_i = 1 \end{cases} \quad (2)$$

$$N_{flag} = \begin{cases} FALSE, & g_n \cap g_t \neq \varnothing \\ TRUE, & g_n \cap g_t = \varnothing \end{cases} \quad (3)$$

*2.2. Match Pair Filter Considering Spatial Overlap Constrain*

After intersection tests, initial match pairs are chose as candidates for image matching. Because footprints of oblique images are larger and more irregular than nadir images, much more image pairs with too narrow or too small overlapped regions would pass intersection tests, which are illustrated in Figure 3. However, feature matching results of these image pairs would not be satisfying in terms of number of matched features and ratio of correct matches because fewer features are existed in the overlapped regions and outlier removal algorithms are prone to failure (Mizotin et al., 2010), e.g., fundamental matrix estimation embedded in the RANSAC (RANdom Sample Consensus) algorithm.

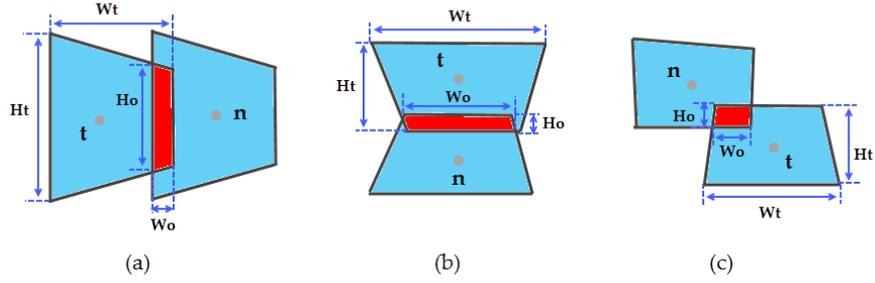

(a)  (b)  (c)

**Figure 3.** Footprints of one image pair. **(a)** and **(b)** indicate that overlapped regions are too narrow; **(c)** stands for too small overlapped region. t and n are the target footprint and the neighboring footprint, respectively; Wt and Ht stand for the width and the height of the envelope of the target footprint, respectively; similarly, Wo and Ho are the width and the height of the overlapped region.

Therefore, initial match pairs are filtered based on SOC in this study. For one match pair, width and height of the overlapped region are calculated, and a proportional value between the overlapped region and the envelope of the target footprint is estimated, as shown in Figure 3. Image pairs would be pruned if the following conditions are not meet as stated in Equation(4), where $W_o$ and $H_o$ are the width and the height of overlapped region, respectively; $W_t$ and $H_t$ are width and height of the target polygon; $R_o$ is an overlap ratio.

$$\begin{cases} W_o \geq W_t \times R_o \\ H_o \geq H_t \times R_o \end{cases} \quad (4)$$

**3. Match Graph Extraction Based on MST-Expansion**

Initial match pairs are obtained after intersection tests based on SDC and SOC constrains. Although the combinatorial complexity of the match pairs is noticeably decreased compared with pairwise matching, vast redundant match pairs are still retained in the initial pair set because only the direct adjacent principle is used for match pair selection. To achieve further simplification of initial match pairs, a non-iterative method for match graph extraction is proposed in this section. This method uses a two-stage algorithm, termed MST-Expansion, to construct a match graph with much fewer image pairs. The algorithm is described in detail in Algorithm 2. First, a topological connection network (Reduced-TCN) is constructed using the match pairs, which is represented by an undirected weighted graph; second, a maximum spanning tree (MST) is extracted from the graph for the simplification of the TCN; finally, the MST is gradually enhanced through local structure analysis in an expansion stage. The three steps are described in the following subsections.

*3.1. Reduced-TCN Construction Considering Overlap Area and Intersection Angle*

To analyze the topological adjacent relationship of the initial match pairs, an undirected weighted graph, termed Reduced-TCN, is constructed, where a vertex stands for one image, and a weighted edge is added for one match pair. The edge weight is used to measure the importance of one match pair. From the aspect of feature matching, matched feature number can be used as a reasonable criterion to quantify the importance of one match pair.

| **Algorithm 1** | Spatial Relationship Constrained Match Pair Selection |
|---|---|

**Input:**  $n$ oblique UAV images, flight control data
**Output:** match pairs $E$ with overlap area and intersection angle

I. **Transform GNSS/IMU data to photogrammetric system**

II. **Calculate images' footprints and center locations**

   1: **for** $j = 0; j < n; j$++ **do**
   2:    Calculate $j^{th}$ image's footprint and store it as $ge_j$
   3:    Calculate $j^{th}$ footprint's center and store it as $gc_j$

III. **Intersection tests considering SDC and SOC**

   1: Build KD-Tree spatial index for point group $G = \{gc_j\}$
   2: **for** each $gc_j$ in $G$ **do**
   3:    Initialize quadrant indicator and non-intersection indicator
   4:    Query $n - 1$ nearest neighbors of $gc_j$ in ascending order $G_{nn} = \{gc_i \mid gc_i \in G, i \neq j\}$
   5:    **for** each $gc_i$ in $G_{nn}$ **do**
   6:     **if** $ge_i$ intersects with $ge_j$ **then**
   7:      Update indicators under the condition that two footprints are overlapped
   8:     **if** overlap region meets SOC constrain **then**
   9:      Calculate area, angle and store them in $E = \{E_{ij} \mid E_{ij} = (area, angle)\}$
 10:    **else**
 11:     Update indicators under the condition that two footprints are not overlapped
 12:    **if** indicators meet termination conditions **then**
 13:     break

IV. **Return match pairs $E$**

In the traditional photogrammetry, platforms always attempt to impose zero pitch and roll angles to obtain nadir images. Usually, the number of matches is positively proportional to the overlap area of one image pair. Therefore, overlap area is suitable enough to measure the importance of one image pair. However, for oblique photogrammetry, imaging angle has dramatic influence on feature matching, which causes varying radiometric and geometric distortion. Almost all feature matching algorithms, including the SIFT algorithm (Lowe, 2004), have limited tolerance to these deformation.

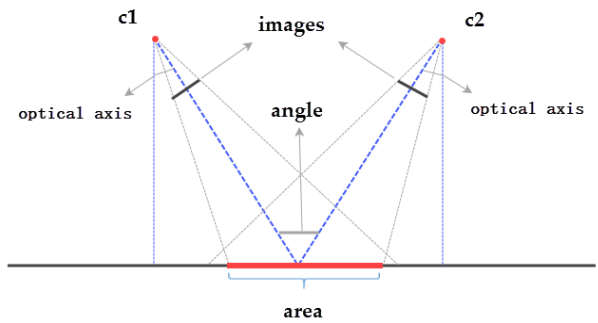

**Figure 4.** Overlap area and intersection angle for edge weight calculation of TCN.

Consequently, in this study, both overlap area and intersection angle are used for the importance calculation of oblique UAV images as shown in Figure 4. The overlap area can be calculated by the Andrew's variant of the Graham scan algorithm (Andrew, 1979), and the intersection angle of one image pair is computed as the angle between two direction vectors, which start from the projection centers to the central locations of footprints.

For the initial match pairs, suppose that $I = \{i_i\}$ and $P = \{p_{ij}\}$ are the image set of size $n$ and the match pair set of size $m$, respectively. Then, the undirected graph $G = (V, E)$ is constructed as follows: define a vertex $v_i$ for each of the $n$ images $i_i$, such that vertex set $V = \{v_i : i = 1,...,n\}$. An undirected edge $e_{ij}$, connecting vertex $v_i$ and vertex $v_j$, is added for each of the $m$ image pairs $p_{ij}$, which forms the edge set $E = \{e_{ij} : i, j = 1,...,n\}$. Additionally, edge weight $w_{ij}$ between a pair of vertices $(v_i, v_j)$ is computed by Equation(5)

$$w_{ij} = R_w \times w_{area} + (1 - R_w) \times w_{angle} \tag{5}$$

where $w_{area}$ and $w_{angle}$ are weights computed from the overlap area and the intersection angle, respectively. These two weight items are calculated by Equation(6) and Equation(7)

$$w_{area} = area / area_{max} \tag{6}$$

$$w_{angle} = \begin{cases} \cos(angle), & angle \leq 90 \\ 0, & otherwise \end{cases} \tag{7}$$

where $area_{max} = \max_{p_{ij} \in P}(area_{p_{ij}})$ is the maximum overlap area. Edge weight $w_{ij}$ is a linear combination of these two weight items, where $R_w$ is the weight ratio ranging from 0.0 to 1.0. Thus, both overlap areas and intersection angles are encoded into the edge weight $w_{ij}$.

*3.2. Maximum Spanning Tree for Reduced-TCN Simplification*

Many redundant edges, deduced from the match pairs, are added to the Reduced-TCN. To bring about the highest efficiency for image matching, the most simplified form of the Reduced-TCN is that all vertices are sequentially chained, which can be constructed by using an MST algorithm. A maximum spanning tree is a subset of edges of an undirected weighted graph, which connects all the vertices together without any cycles and with the maximum total edge weight. When the importance of one image pair is assigned as the edge weight, the MST, extracted from the Reduced-TCN, can be considered as an image concatenation queue with the least match pairs and with the most stability than the other orders.

In this study, the Kruskal's algorithm (Kruskal, 1956) is used to extract the MST from the Reduced-TCN. The primal algorithm is aimed to search a minimum spanning tree, which is the same as the maximum spanning tree algorithm except that the minimum total edge weight is gathered. Therefore, the values of edge weights are negated before applying the algorithm on the Reduced-TCN.

*3.3. MST expansion for Connection Network Enhancement*

After simplification based on the MST algorithm, a vast majority of vertices in the MST just have connections in the spanning direction. To strengthen image connection network, the MST-Expansion algorithm expands critical connections in the direction orthogonal to the spanning direction, which is mainly dependent on local structure analysis. In this study, the direction orthogonal to the spanning direction is termed expansion direction. The detailed processing for MST expansion is described and illustrated as follows.

Suppose that the Reduced-TCN is represented by an undirected graph $g\_tcn$, a new TCN, represented by a graph $g\_mst\_expansion$, is firstly created. This graph establishes spatial adjacent relationship of the simplified Reduced-TCN and initialized from the MST. Hence, $g\_mst\_expansion$ is a subgraph of $g\_tcn$, as shown in Figure 5(a), where black circles and solid lines stand for vertices and edges of graph $g\_mst\_expansion$, respectively. The edges of graph $g\_tcn$ are indicated by both dashed and solid lines. Then, for each vertex in graph $g\_mst\_expansion$, an expansion operation is conducted, which consists of four major steps: (1) determine the expansion direction in $g\_mst\_expansion$; (2) search existed edge connections in $g\_mst\_expansion$; (3) search expansion candidates in $g\_tcn$; (4) execute candidate expansion in $g\_mst\_expansion$.

For the first step of the expansion operation, the expansion direction is determined based on eigenvector analysis of a covariance matrix. Incident neighbors of current vertex are searched from graph $g\_mst\_expansion$, as shown in Figure 5(b)-(e), where the red circle is the current vertex and neighboring vertices are rendered in green color. Then, a covariance matrix is computed using ground coordinates of the vertex set, which consists of the current vertex and its neighboring vertices. Based on SVD decomposition of the covariance matrix, the spanning and expansion directions can be simultaneously determined by the eigenvectors corresponding to the maximum and minimum eigenvalues, respectively. Considering that eigenvalues indicate spatial distribution of involved vertices, the expansion operation would be skipped if the eigenvalue ratio is less than a specified threshold as shown by Equation(8), where $ev_{max}$ and $ev_{min}$ are the maximum and minimum eigenvalues, respectively; $R_e$ is the threshold of eigenvalue ratio.

$$ev_{max} \leq R_e \times ev_{min} \tag{8}$$

The second step is to search existed connections in expansion regions. Two direction vectors are firstly generated by rotating the expansion vector by an angle $\alpha$ clockwise and counter-clockwise. Then, two regions, in which the expansion vector locates, are used to define the expansion regions, which restricts candidate searching. As shown in Figure 5(b)-(e), dashed blue lines represent boundaries of these two expansion regions. Meanwhile, existed connections in the expansion regions are searched and counted from $g\_mst\_expansion$. When the number of existed connections in both of these two expansion regions is greater than or equal to the expansion threshold $T_e$, the expansion operation would be skipped for current vertex because stable edge connections are established for this vertex.

The third step is to search candidates in expansion regions. A candidate is a vertex that locates in expansion regions and that has connected to current vertex in graph $g\_tcn$, but not in $g\_mst\_expansion$. Thus, Incident neighbors of current vertex are searched from graph $g\_tcn$, and candidates are filtered by checking whether or not the edge, connecting it to current vertex, is existed in graph $g\_mst\_expansion$. If not, the vertex would be labeled as one candidate. As shown in Figure 5(b)-(e), all candidates are rendered in blue circles and connected by dashed lines, which represent edges in graph $g\_tcn$. There are two sets of candidates corresponding to two expansion regions. For each set, candidates are sorted in descending order by edge weight.

The fourth step is to expand image connections in two expansion regions. For each expansion region, the edge, connecting one candidate to current vertex, is added to graph $g\_mst\_expansion$ until expansion count reaches the specified expansion threshold. After the expansion operation has been executed on each vertex, a more stable image connection network is constructed as presented in Figure 5(f). In this example, the expansion threshold is set as one, and there are four extra connections have been added to the final image connection network. Comparing with the initial MST graph as shown in Figure 5(a), the MST-Expansion algorithm can enhance the stability of the image connection network and generate the match graph, as shown in Figure 5(f), to guide image matching.

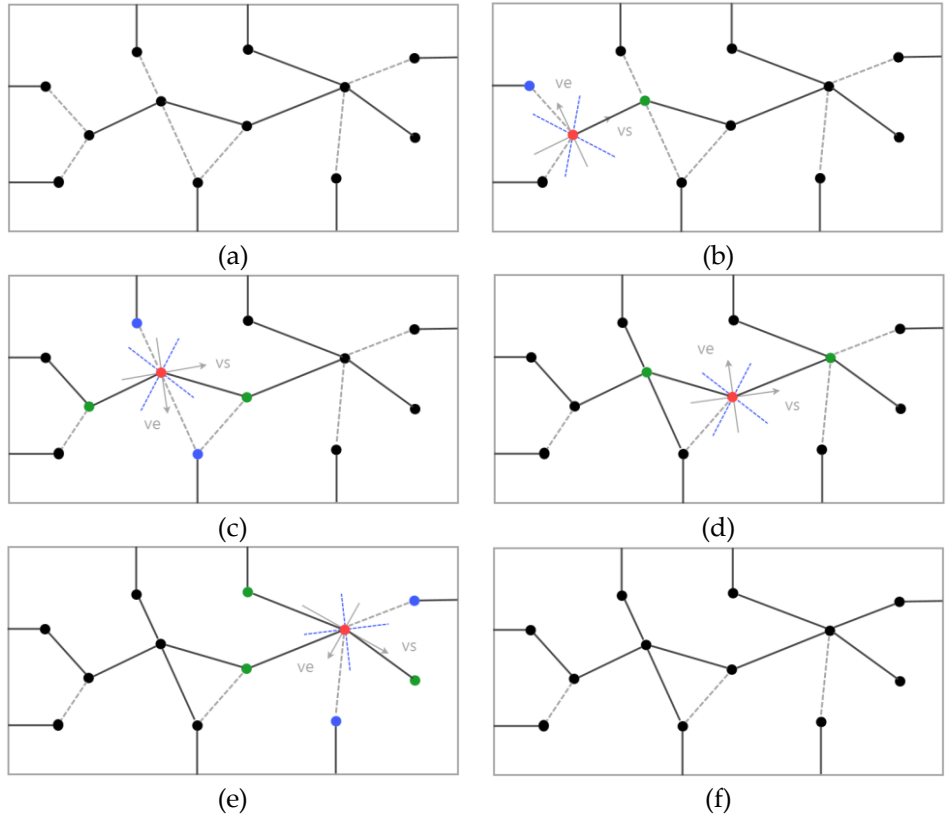

**Figure 5.** Illustration of MST-Expansion algorithm for enhancement of image connection network. **(a)** is the initial match graph; **(b)**, **(c)**, **(d)** and **(e)** are the results of an expansion operation on four vertices; **(d)** is the expanded match graph. Black circles stand for vertices. The red circle stands for the current vertex of an expansion operation. Green circles indicate incident neighbors. Blue circles indicate the candidates of the expansion. Blue lines in dash style restrict the expansion regions. The spanning and expansion directions are indicated by two vectors in gray color, named vs and ve, respectively.

## 4. Efficient SfM Guided by Match Graph

The efficient SfM guided by match graph mainly consists of three parts. The SiftGPU open-source software package (Wu, 2007), a GPU (Graphic Process Unit) implementation of the SIFT (Scale Invariant Feature Transform) algorithm, is firstly adopted to achieve fast SIFT feature extraction. To process large images, SiftGPU utilizes the down-sampling strategy to reduce memory storage costs. Because of very high spatial resolution of UAV images, the default configuration of SiftGPU is used in this study without too many worries about the number of extracted features.

Extracted features then can be matched by comparing two sets of feature descriptors, in which correspondences can be identified when the ratio test is satisfying (Lowe, 2004). Instead of pairwise matching, only image pairs corresponding to edges in the match graph are used for feature matching. To remove outliers, the RANSAC algorithm (Fischler and Bolles, 1981), embedded with an epipolar geometric estimation based on a fundamental matrix, is applied in this paper. To recover camera poses and scene geometry, consistent correspondences from multiple image pairs are tied to generate tracks, which are a connected set of points across multiple images, corresponding to the same 3D point in the scene.

The problem for recovering camera poses and scene geometry can then be formulated as a joint minimization of a non-linear function, where the sum of errors between projections of tracks and the corresponding image points is minimized as presented by Equation(9)

$$\min_{C_j, X_i} \sum_{i=1}^{n} \sum_{j=1}^{m} \rho_{ij} \left\| P(C_j, X_i) - x_{ij} \right\|^2 \tag{9}$$

where $X_i$ and $C_j$ indicate a 3D point and a camera, respectively; $P(C_j, X_i)$ is the predicted projection of point $X_i$ on camera $C_j$; $x_{ij}$ is an observed image point; $\|\cdot\|$ denotes L$^2$-norm; $\rho_{ij}$ is an indicator function with $\rho_{ij} = 1$ if point $X_i$ is visible in camera $C_j$; otherwise $\rho_{ij} = 0$.

The minimization problem is solved using the open-source optimization software Ceres Solver (Sameer and Keir, 2010), which facilitates the problem solving with automatic differential and convenient methods for cost function definition. Because good initial estimates for unknown parameters are essential to ensure a global optimal solution, a standard incremental SfM pipeline, similar to (Snavely et al., 2006), is used rather than estimating the parameters for all cameras and 3D points at once. In the incremental estimation procedure, a local bundle adjustment is performed to reduce accumulative errors among local image connections when the count of new registered images reaches a specified number. After all images have been registered, a global bundle adjustment is conducted to refine all parameters. Camera poses and 3D points would be estimated in an arbitrary system. In this study, GCPs (Ground Control Points), is utilized to achieve absolute geo-referencing by adding extra constrains to the joint minimization problem, whose objective function is similar to Equation(9) except that the 3D points is kept constant during the iterative optimization procedure.

| | |
|---|---|
| **Algorithm 2** | Match Graph Extraction Based on MST-Expansion |

**Input:**   match pairs $E$ with overlap area and intersection angle
**Output:**   match graph $g\_mst\_expansion$

I. **Topological connection network construction**
  1: Create a new graph $g\_tcn$ with $sizeof(E)$ vertices
  2: **for** each $e$ in $E$ **do**
  3:    Add a new edge with weight of $weight(area, angle)$ to $g\_tcn$

II. **Maximum spanning tree extraction for TCN simplification**
  1: Extract the maximum spanning tree $mst$ from $g\_tcn$
  2: Create a new graph $g\_mst\_expansion$ corresponding to $mst$

III. **MST expansion for connection network enhancement**
  1: **for** each vertex $v$ in $vertext(g\_mst\_expansion)$ **do**
  2:    Search neighboring vertex set $V_1$ of $v$ in graph $g\_mst\_expansion$
  3:    Eigen analysis for covariance matrix of $V_1$, $(ev_{max}, ev_{min}) = eigen(V_1)$
  4:    **if** $ev_{max}/ev_{min}$ is less than $R_e$ **then**
  5:      **continue**
  6:    Calculate expansion region $g1$ and $g2$
  7:    Count number of vertices in expansion region $(n1, n2) = count(g1, g2, V_1)$
  8:    **if** $n1$ and $n2$ greater than or equal to $T_e$ **then**
  9:      **continue**
  10:   Search neighboring vertex set $V_2$ of $v$ in graph $g\_tcn$
  11:   Query candidates $C_1$ and $C_2$ in expansion region $g1$ and $g2$ from $V_2$
  12:   Sort candidates $C_1$ and $C_2$ by weight in ascending order
  13:   **if** $n1$ less than $T_e$ **then**
  14:      Expansion with candidates $C_1$ until $n1$ greater than or equal to $T_e$
  15:   **if** $n2$ less than $T_e$ **then**
  16:      Expansion with candidates $C_2$ until $n2$ greater than or equal to $T_e$

IV. **Return match graph $g\_mst\_expansion$**

## 5. Experiment and Results

In the experiments, three datasets are used to check the validation of the proposed algorithms. First, we analyze the efficiency achieved from SDC and search the optimal overlay ratio for SOC used in the SRC-InterTest algorithm. Meanwhile, the weight ratio between overlap area and intersection angle is also analyzed for its influence on the stability of the match graph. Then, match graphs are constructed using the MST-Expansion algorithms. In addition, four methods, including Full-TCN, Reduced-TCN, MST and MST-Expansion, are compared in terms of efficiency, completeness and accuracy. Finally, a comparison between MST-Expansion and MicMac (Deseilligny and Clery, 2011) are performed.

In the following tests, the same set of parameters is adopted with non-intersection count $T_I=8$, eigenvalue ratio $R_e=3$, expansion angle $\alpha=45°$ and expansion threshold $T_e=1$. For photogrammetric measurement, all cameras are calibrated using a model with 8 parameters, including one for focal length, two for principle point, three and two for coefficients of radial distortion and tangent distortion, respectively. For system calibration, the lever-arm[1] offset is assumed to be zero because the distance to the scene is much larger than the lever-arm offset. It is too hard to measure the bore-sight[2] angles because the IMU sensor is integrated into the UAV platform. Therefore, only camera mounting angles are used to approximate the relative poses between UAV platforms and camera sensors.

*5.1. Test Sites and Datasets*

The detailed information for flight configuration is listed in Table 1. For the three test sites, a multi-rotor UAV platform, equipped with three different oblique photogrammetric systems, has been used for the campaigns of data acquisition. The first test site locates in a suburb area, where some railroads come across. This region is mainly covered by vegetation. For this test site, an imaging instrument with one Sony RX1R camera has been integrated with the UAV platform, which is mounted with a pitch angle of 25° and a roll angle of -15°. Under the flight height of 165 m relative to the position from which the UAV takes off, 320 images, dimensions of 6000 by 4000 pixels, are acquired and the GSD (Ground Sampling Distance) is about 5.05 cm. Therefore, overlap degrees in forward and side directions are 85% and 45%, respectively. The ground coverage of each image is presented in Figure 7(a).

Table 1. Detailed information for flight configuration of the three test sites.

| Item Name | Test Site 1 | Test Site 2 | Test Site 3 |
| --- | --- | --- | --- |
| UAV type | multi-rotor | multi-rotor | multi-rotor |
| Flight height (m) | 165 | 120 | 175 |
| Forward/side overlap (%) | 85/45 | 85/50 | 75/70 |
| Camera mode | Sony RX1R | Sony RX1R | Sony NEX-7 |
| Number of cameras | 1 | 2 | 5 |
| Sensor size (mm×mm) | 35.8×23.9 | 35.8×23.9 | 23.4 × 15.6 |
| Focal length (mm) | 35 | 35 | nadir: 16 oblique: 35 |
| Camera mount angle (°) | front: 25, -15 | front: 25, -15 back: 0, -25 | nadir: 0 oblique: 45/-45 |
| Number of images | 320 | 390 | 750 |
| Image size (pixel×pixel) | 6000×4000 | 6000×4000 | 6000×4000 |
| GSD (cm) | 5.05 | 3.67 | 4.27 |

The second test site is covered by farmlands in which several roads exist. In this test site, 390 images are captured by the same camera as one used in the first site while a dual-camera oblique photogrammetric system, consists of a front camera and a back camera, has been designed and integrated for data acquisition. This dual-camera imaging system could simultaneously record images in both forward and side directions. In this study, the front camera is mounted with 25° and -15° for pitch and roll angles, respectively. The back camera is designed to capture images in across-track direction. Then, a roll angle of -25° has been configured for the back camera. Under flight height of 120 m, the GSD is about 3.67 cm, and forward and side overlap degrees are 85% and 50%, respectively. The ground coverage of each camera is shown in Figure 7(b). The detailed design of the oblique photogrammetric system used in test site 1 and test site 2 is documented in (Jiang et al., 2017), except that only the front camera is used in test site 1. In addition, 43 GCPs have been selected for the assessment of geo-referencing accuracy, which are uniformly distributed in this test site as shown by Figure 6. Accurate coordinate measurements are performed with a Trimble R8 GNSS receiver, for which horizontal and vertical accuracies are 1 cm and 2 cm in RTK-GNSS (Real Time Kinematic GNSS) mode, respectively.

---

[1] The offset between the phase center of a GNSS antenna and the optical center of a camera sensor.

[2] The rotational angles between the axes of an IMU coordinate system and a camera coordinate system.

In the field of photogrammetry, oblique systems integrated with a five-camera imaging instrument have been commonly adopted for data acquisition in urban regions. Therefore, an urban region is selected as the third test site, where a shopping plaza locates in the central region and be surrounded by some residential buildings. In this study, images from the third test site are captured by a conventional five-camera oblique photogrammetric system, for which one nadir camera and four oblique cameras are integrated. Four oblique cameras are rotated by 45° with respect to the nadir camera. The Sony NEX-7 camera with dimensions of 6000 by 4000 pixels has been selected. Under flight height of 175 m, there are 750 total images captured. Overlap degrees in forward and side directions are measured for nadir images, and values of 75% and 70% are achieved. The final GSD of nadir images is about 4.27 cm for this dataset. The ground coverage of this dataset is shown in Figure 7(c).

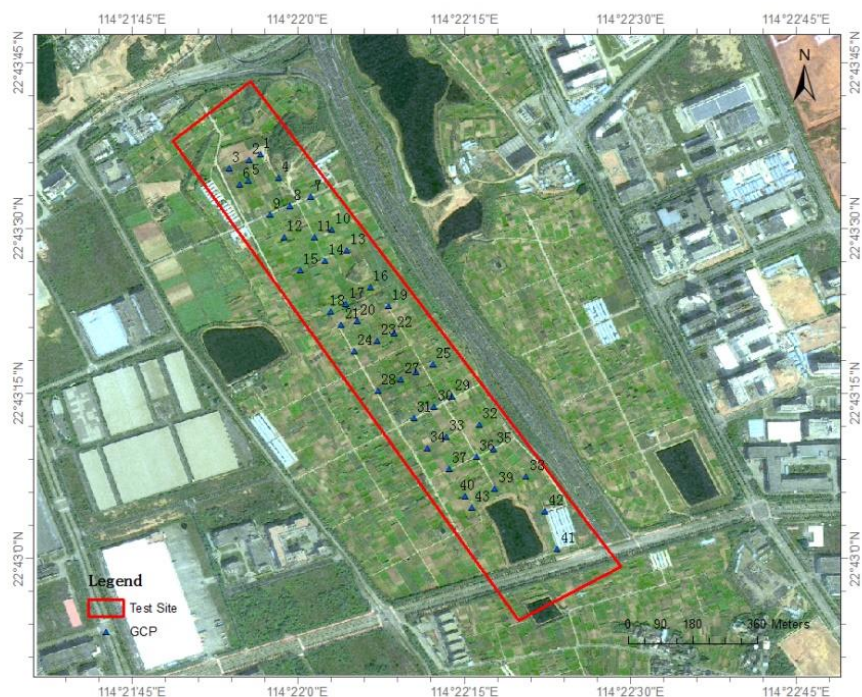

**Figure 6.** GCP distribution in the second test site. The red polygon stands for approximate test region. GCPs are rendered as blue triangles.

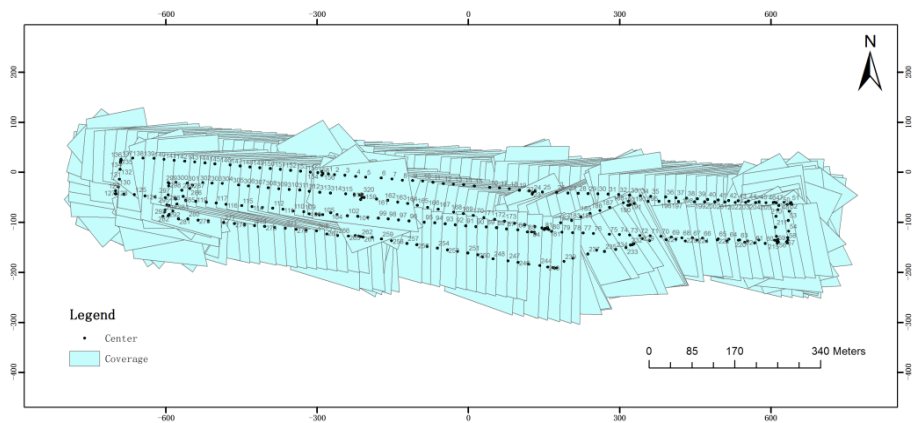

(a)

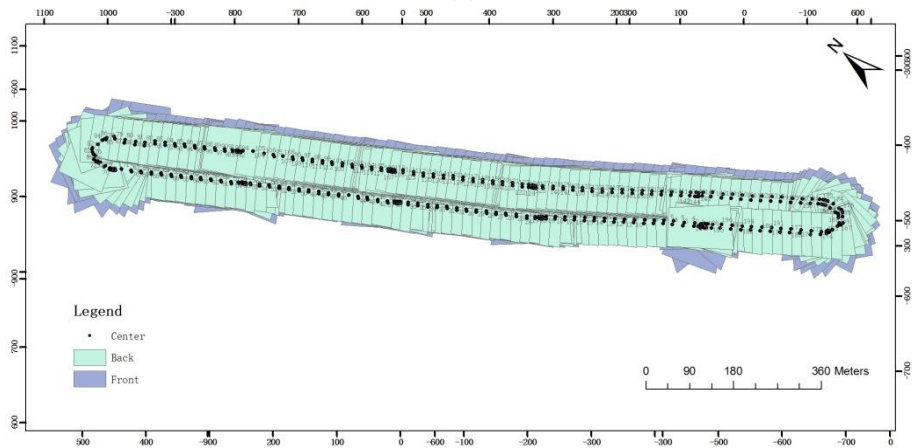

(b)

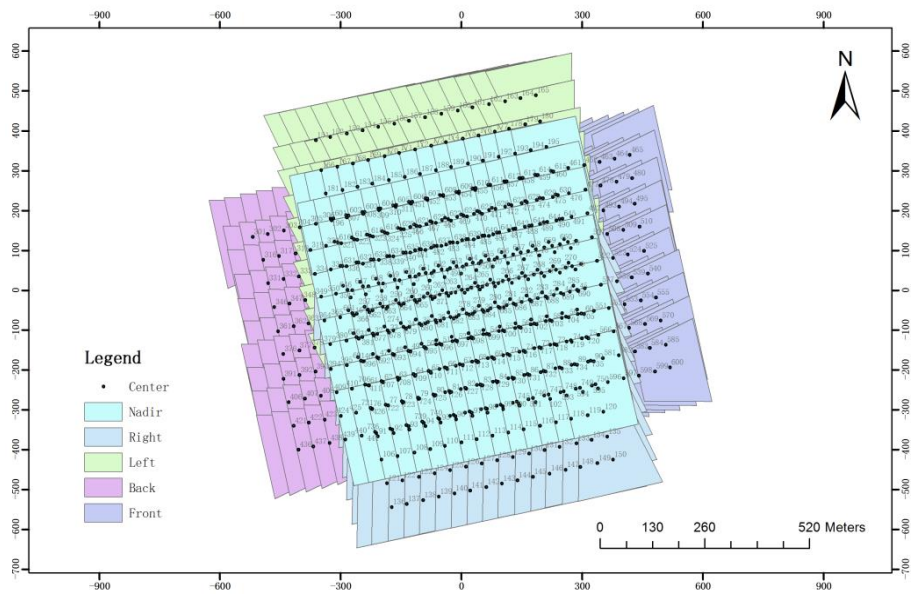

(c)

**Figure 7.** Ground coverage of the three datasets. **(a)** ground coverage of the first dataset; **(b)** ground coverage of the second dataset; **(c)** ground coverage of the third dataset.

*5.2. Analysis Influence of SDC and SOC on Match Pair Selection*

In the SRC-InterTest algorithm, spatial relationship constrains, including SDC and SOC, are used for fast and reliable intersection tests, which provide initial match pairs to construct the TCN graph. SDC is independent of a distance threshold for efficient intersection tests, and SOC is used to filter non-essential match pairs to not only simplify TCN construction, but also decrease ratio of erroneous matches. Thus, SRC-InterTest plays a crucial role in the pipeline of the efficient SfM. In this section, we will analyze what efficiency is achieved from SDC and what overlay ratio is optimal for SOC. In this test, the weight ratio $R_w = 0.6$.

To analyze the efficiency achieved from SDC, intersection tests with and without SDC, are conducted and number of tests is used as the criterion for efficiency assessment. Number of tests for each test site is presented by Figure 8, where the total number of tests is shown by Figure 8(a) and the mean number of tests is presented by Figure 8(b). The results show that total number of tests is quadratic in image number when SDC is absent, which could be observed from the approximate parabola curve in Figure 8(a). On the contrary, SDC could dramatically decrease the total number of tests and guarantee that the mean number of tests for each image is not of correlation to image number. Thus, a near constant number of intersection tests are needed when SDC is adopted, which can be deduced from the almost horizontal curve in Figure 8(b). The curve is not strictly horizontal due to the fact that larger oblique angles used in the third dataset lead to much more intersected image pairs. Thus, SDC could enable intersection tests for image pair selection with linear time complexity.

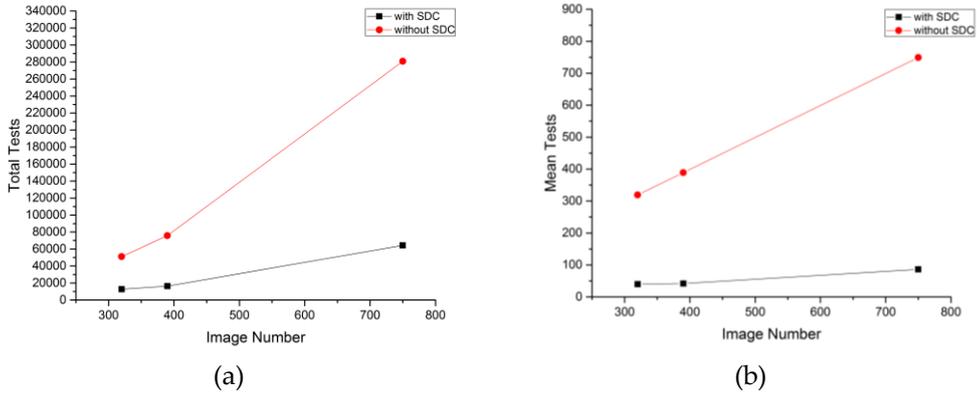

(a)          (b)

**Figure 8.** Number of intersection tests with and without SDC. **(a)** Total number of tests for the three datasets; **(b)** Mean number of tests for the three test datasets.

To analyze the influence of the overlay ratio, the overlay ratios in SOC are uniformly sampled between 0.0 and 0.8 with interval value 0.1. Number of connected images and RMSE (Root Mean Square Error) estimated by bundle adjustment are set as the criteria to assess overlay ratio. The results are shown in Figure 9, where the influence of the overlay ratio on number of connected images is presented by Figure 9(a) and the influence on RMSE is illustrated by Figure 9(b). Then, some conclusions could be made. First, all images could be successfully connected when overlay ratio is less than 0.5, and the number of connected images is dramatically decreased when overlay ratio is greater than 0.5, especially for site 2 and site 3. Second, no remarkable change on RMSE is observed when overlay ratio is less than 0.5. However, RMSEs for the three datasets rapidly descended when overlay ratio is greater than 0.6, which could be explained by the dramatic decrease of connected images. Therefore, the overlay ratio is set as 0.5 in this paper to both filter non-essential image pairs and achieve the best reconstruction completeness. The results of image pair selection for the three datasets are shown in Figure 10, where image pairs are represented by adjacent matrixes. Figure 10(a) – (c) stand for image pair selection results without SOC, and Figure 10(d) – (f) are the selection results with overlay ratio set as 0.5. Under the SOC constrain, image pairs with too narrow or too small overlay regions are pruned out, and near one second image pairs are preserved comparing with image pair selection without the SOC constrain.

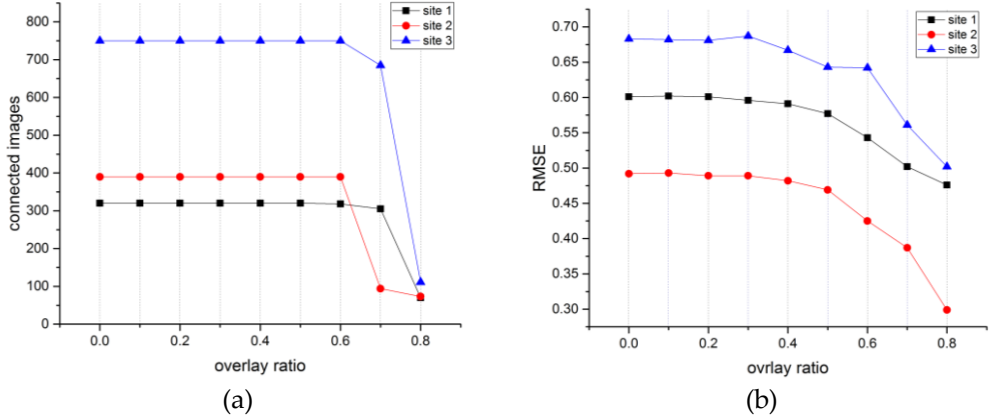

**Figure 9.** The influence of the overlay ratio on number of connected images and RMSE estimated by bundle adjustment. **(a)** influence on number of connected images. **(b)** influence on RMSE.

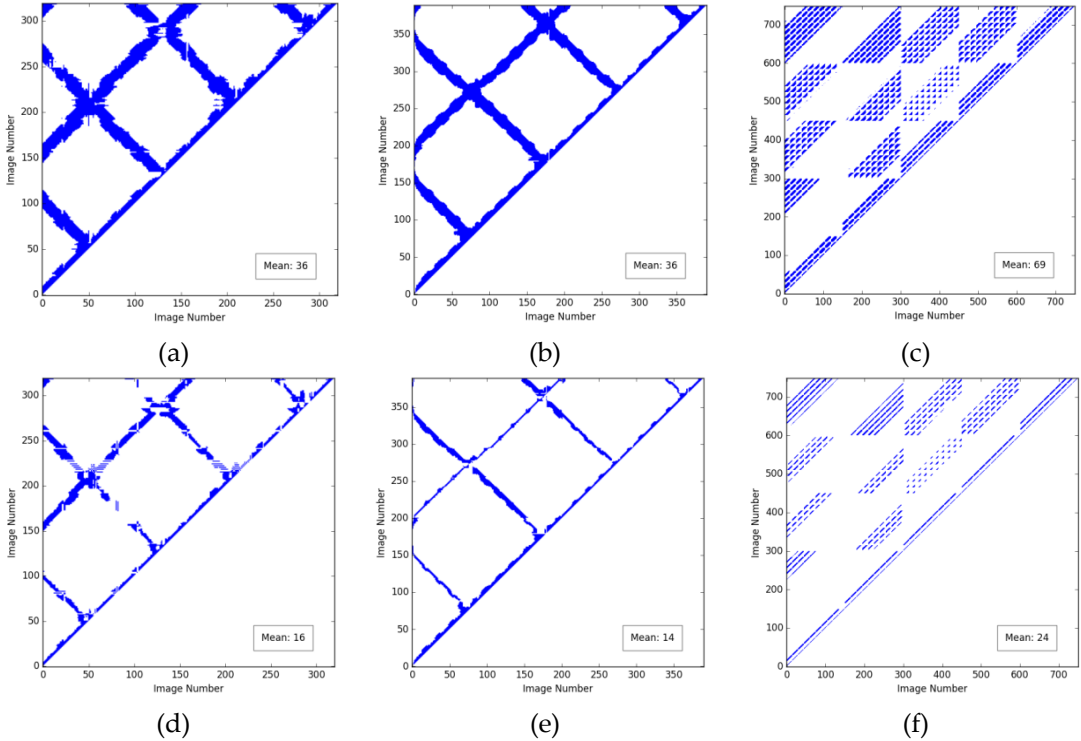

**Figure 10.** Matrix representation of match pair selection results of the three datasets. **(a)**-**(c)** full adjacent matrix. **(d)**-**(f)** reduced adjacent matrix. The mean value near the bottom right corner indicates the average number of connections for each image.

*5.3. Analysis Influence of Overlap Area and Intersection Angle on Edge Weighting*

In TCN construction, edge weight is used to measure the importance of one match pair, which is quantified by the number of matched features in this paper. The match pair with larger number of matched features would be assigned a higher weight; otherwise, a lower weight would be given. In the traditional photogrammetry, nadir images with near zero pitch and roll angles are usually captured, and the number of matched features would be positive correlative to overlap area. However, for oblique images, the number of matched features is noticeably influenced by imaging angles. Thus, both overlap area and intersection angle are used to calculate edge weight for TCN. In this test, the overlap ratio $R_o = 0.5$.

To analyze the influence of weight ratio between overlap area and intersection angle, dataset 3 is selected for bundle adjustment experiments because varying configurations for overlap area and imaging angle are existed. Weight ratios ranging from 0.0 to 1.0 are evenly sampled with interval 0.1, and two criterions, including the number of connected images and RMSE estimated from bundle adjustment, are used for evaluation as shown in Figure 11. Experimental results show that all images could be connected when weight ratio is valued from 0.4 to 0.6. However, 700 images are oriented when only overlap area is used for edge weight calculation because weight ratio is 1.0. The reason for this result is that the match pair captured in the opposite direction has larger overlap area and a larger weight is assigned for the corresponding edge when just considering overlap area. Then, the edge has high priority to construct match graph. However, much less features could be matched from the match pair, which leads to an unstable image connection network. Thus, more and more unstable edges are involved in the construction of match graph when weight ratio is increased from 0.7 to 1.0, which could be proved by the gradually decreased number of connection images as shown in Figure 11(a).

In addition, when only intersection angle is used for edge weight calculation, merely 550 images are connected because overlapped regions could not be guaranteed only through the analysis of intersection angle. However, the number of connected images is rapidly increased with the increase of weight ratio from 0.1 to 0.3. The main reason is that much more overlapped match pairs participate in generation of match graph. In this paper, the weight ratio is set as 0.6 because the lowest RMSE has been achieved as illustrated by Figure 11(b).

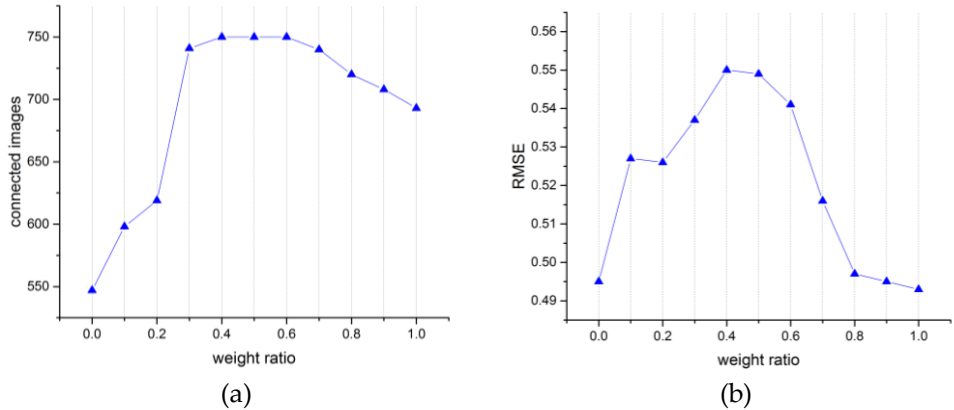

(a)    (b)

**Figure 11.** The influence of weight ratio on the number of connected images and RMSE estimated by bundle adjustment from test site 3. **(a)** influence on the number of connected images. **(b)** influence on RMSE of bundle adjustment.

*5.4. TCN Construction and Match Graph Extraction*

TCN is represented by an undirected weighted graph and could be constructed based on selected match pairs and computed edge weights. In this paper, the TCN is represented by a weight matrix, in which a non-zero element stands for one match pair and its value indicates the corresponding edge weight. Matrix representation of the Full-TCN is illustrated by Figure 12, where Figure 12(a), Figure 12(b) and Figure 12(c) are weight matrixes of dataset 1, dataset 2 and dataset 3, respectively. The Full-TCN is the graph which contains all intersected match pairs without SOC constrain. We can see that some match pairs with low weights are existed in these three matrixes, which are rendered in green color. Usually, these match pairs have smaller overlap area or larger intersection angle, and fewer or no features can be matched.

Therefore, the Reduce-TCN is generated under SOC constrain, which filters a majority of match pairs with low weights. Similarly, matrix representation of the Reduced-TCN is shown in Figure 13. For the three datasets, a larger weight is assigned for the edge connecting two adjacent images, which can be observed from diagonal values of each weight matrix. This can be explained by the larger overlap area and the smaller intersection angle of adjacent images. Additionally, dataset 1 and dataset 2 have almost the same matrix structure and element value except that in dataset 1, larger weights are calculated for the edges connecting images from adjacent tracks because these match pairs have larger overlap area. For dataset 3, the match pairs consisted of nadir images have the smallest intersection angles, which leads to the largest weights assigned as depicted in the top-right part of Figure 13(c). Meanwhile, the edges connecting a nadir image and an oblique image have relative larger weights than the edges connecting two oblique images as illustrated in the top part of Figure 13(c). Compared with the Full-TCN, much fewer edges with low weights exist and the total number of edges is noticeable decreased in the Reduce-TCN.

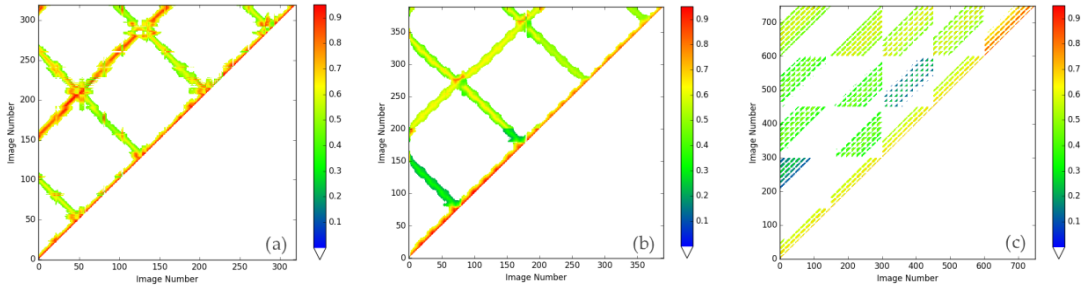

**Figure 12.** Weight matrix representation of Full-TCN for each dataset. **(a)** weight matrix for dataset 1. **(b)** weight matrix dataset 2. **(c)** weight matrix for dataset 3.

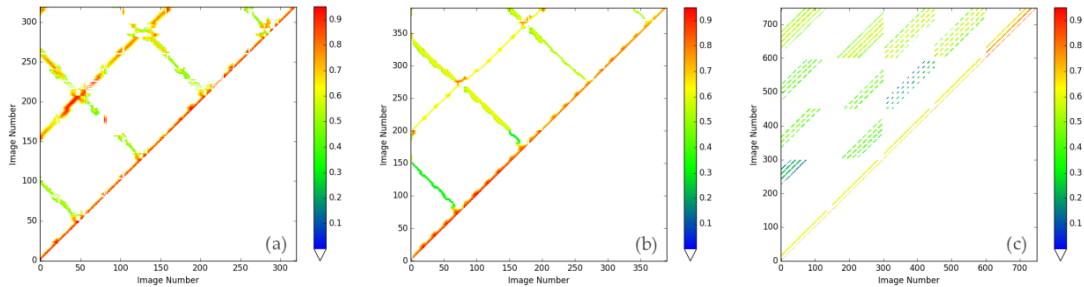

**Figure 13.** Weight matrix representation of Reduced-TCN for each dataset. **(a)** weight matrix for dataset 1. **(b)** weight matrix dataset 2. **(c)** weight matrix for dataset 3.

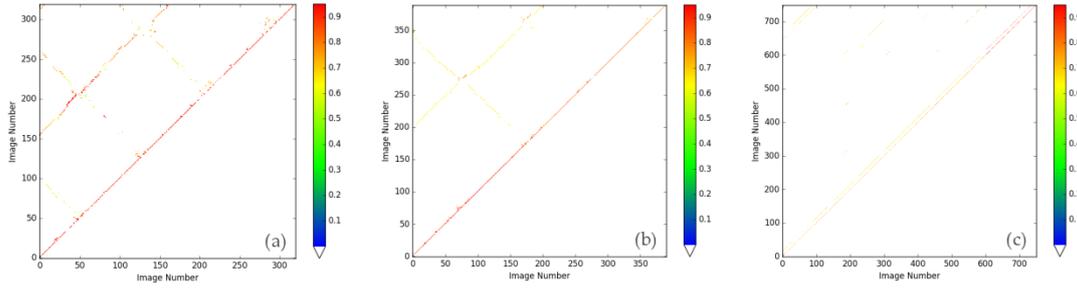

**Figure 14.** Weight matrix representation of match graph for each dataset. **(a)** weight matrix for dataset 1. **(b)** weight matrix for dataset 2. **(c)** weight matrix for dataset 3.

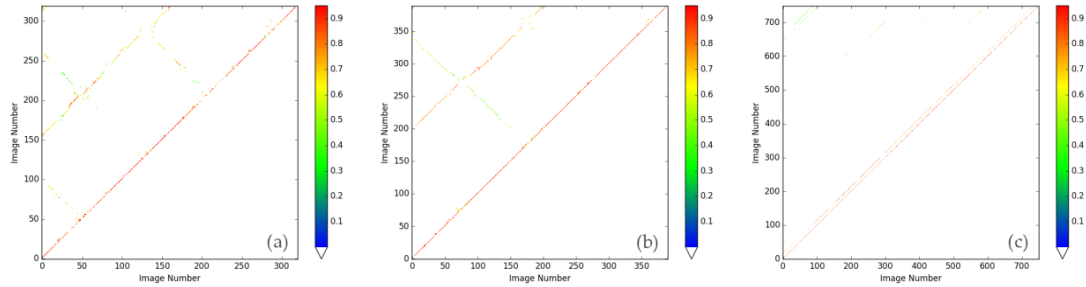

**Figure 15.** Weight matrix representation of feature number for each dataset. **(a)** weight matrix for dataset 1. **(b)** weight matrix for dataset 2. **(c)** weight matrix for dataset 3.

Figure 14 shows the results of match graph extraction. First, the initial Reduce-TCN is simplified based on MST extraction, which links all images using fewer edges and preserving as much stability as possible. Then, the MST is expanded to enhance the image connection network. Observing the matrix representation of the match graph, we can see that almost all edges along track have been extracted, which could be verified by the diagonal elements in each matrix. Because of large overlap area and small intersection angle, these edges are assigned with high weights. For dataset 1, match pairs in both the same and the opposite directions have been selected as depicted in the top-left part of the weight matrix. However, for dataset 2, only match pairs in the same direction are selected for each camera as shown in both the bottom-left and the top-right parts of Figure 14(b). The main reason is that the MST-Expansion algorithm enhances image connection network by adding edges connecting two images from front camera and back camera, respectively, due to the relative smaller intersection angle. For dataset 3, except for edges connecting two adjacent images along track, almost all the other edges are preserved to connect one nadir image and one oblique image. These edges are crucial to stabilize image connection network. To verify correction of edge weights, feature matching is executed guided by the match graph and results are represented in Figure 15, which shows that match matrixes are almost consistent with weight matrixes.

*5.5. Comparison of Efficiency, Completeness and Accuracy*

In order to assess the MST-Expansion algorithm, the comparison in terms of efficiency, completeness and accuracy is conducted in this section. First, efficiency is evaluated by the number of match pairs involved in feature matching and the processing time of feature matching. Second, the number of reconstructed images and 3D points are commonly used to assess the completeness for reconstruction. However, in this study, more attention has been paid to the number of oriented images because 3D points can be resumed by post-processing, e.g., dense matching. Finally, RMSE estimated from bundle adjustment is used for accuracy comparison. To evaluate the geo-referencing accuracy, the GCPs in the second dataset are utilized to achieve absolute orientation. Besides, accuracy comparison with the open-source software package MicMac is also conducted. Comparison between Full-TCN, Reduced-TCN, MST and MST-Expansion is considered in this study. All experiments are executed on an Intel Core i7-4710HQ laptop on the Windows platform with a 2.5 GHz CPU and a 2.0 G GeForce GTX 860M graphics card.

5.5.1. Efficiency

The number of match pairs and feature matching time are used for assessment and analysis of efficiency. Match pairs are firstly extracted from the corresponding match graph for each dataset, where each edge indicates one match pair. Statistic results for the number of match pairs and the processing time of feature matching are shown in Figure 16(a) and Figure 16(b), respectively. Because of failed reconstruction, the MST is not used in the comparison.

It is clearly show that much fewer match pairs are retained from the MST-Expansion algorithm, whose value is 640, 664 and 1516 for dataset 1, dataset 2 and dataset 3, respectively. However, the Full-TCNs have the largest number of match pairs for the three datasets with value 10239, 12384 and 52248, respectively. Although more than one half match pairs with too narrow or too small overlap area are pruned under the SOC constrain, the Reduced-TCNs have exceeded eight times number of match pairs than the MST-Expansions. Noticeably, for dataset 3, the ratio of the match pair number between Full-TCN and MST-Expansion is about 35, while the ratio for dataset 1 and dataset 2 is no more than 19. The main reason is that oblique cameras have been mounted with larger angles in dataset 3 and the overlap degree is higher than the other two datasets, which would generate much more intersected match pairs. However, a large proportion of the match pairs is not satisfying under the SOC and would be filtered out because of irregular ground coverage geometry. Additionally, the Reduced-TCN will be simplified by the MST-Expansion algorithm. Thus, higher efficiency could be achieved with the increase of image overlap degree.

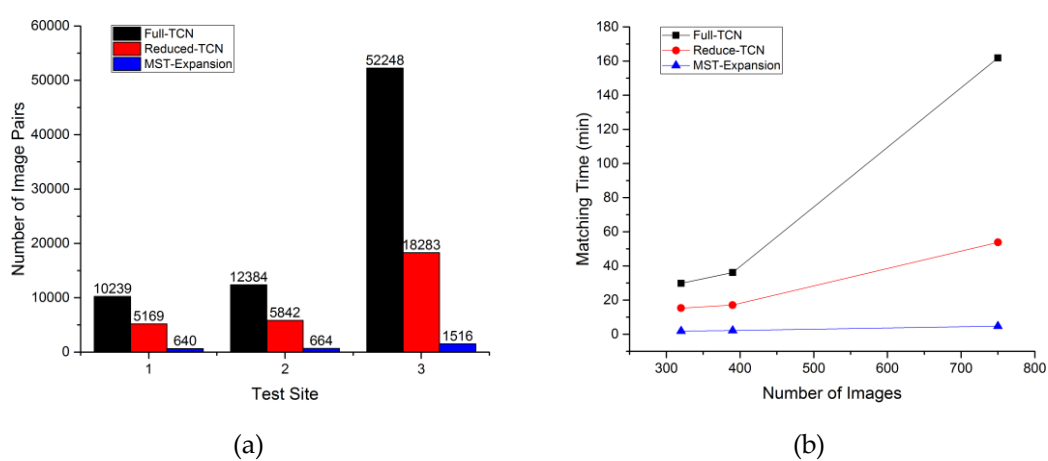

(a) (b)

**Figure 16.** Number of image pairs and feature matching time for efficiency comparison. **(a)** number of image pairs for efficiency comparison. **(b)** feature matching time for efficiency comparison (time in minute).

Figure 16(b) fits the curves of feature matching time over the number of images for the three datasets, where the number of images is 320, 390 and 750 for dataset 1, dataset 2 and dataset 3, respectively. The unit of feature matching time is in minute. The results indicate that feature matching time for both the Full-TCN and the Reduced-TCN is quadratic in the number of images, which can be verified through the fitted parabolic curves. On the contrary, time complexity for the MST-Expansion is almost linear in the image number, for which time consumption for the three datasets is about 1.9 min, 2.2 min and 4.8 min, respectively. Therefore, the MST-Expansion has the lowest time complexity for feature matching.

To visually analyze the efficiency of different methods, image connection networks are constructed and illustrated in Figure 17, where a node stands for one image and an edge indicates two intersected images. Nodes in the network are rendered by red circles and edges are drawn by gray lines. Match graphs of Full-TCN, Reduced-TCN, MST and MST-Expansion are represented by Figure 17(a), Figure 17(b), Figure 17(c) and Figure 17(d), respectively. It is clearly shown that compared with the Full-TCN, edge density in the Reduced-TCN is dramatically decreased under SOC constrain, which is further simplified through MST. Vast majority of nodes have two incident neighbors in the forward direction by checking Figure 17(c). On the contrary, through expansion, almost all nodes are connected to others in the directions of both along-track and across-track, which generates a more stable image connection network as presented in Figure 17(d). Match graphs for the other two datasets are presented in Figure A1.

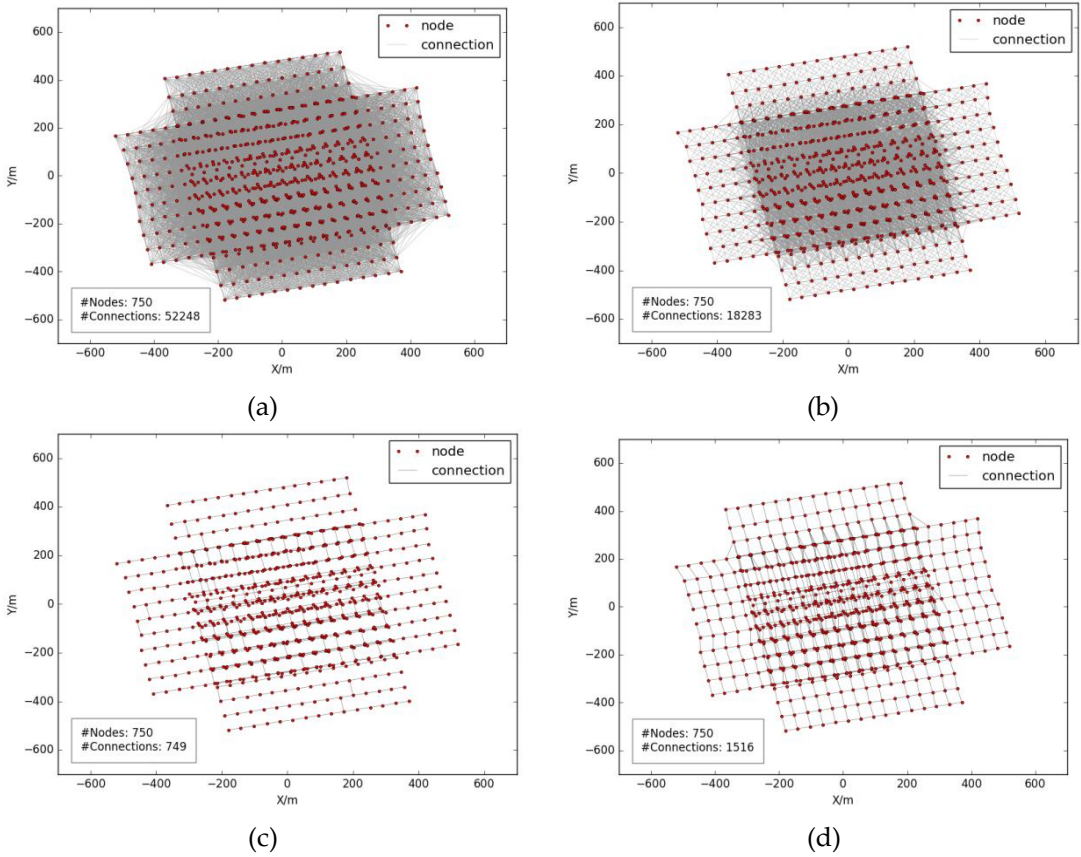

**Figure 17.** Image connection network for dataset 3. **(a)** Full-TCN network. **(b)** Reduced-TCN network. **(c)** MST network. **(d)** MST-Expansion network.

5.5.2. Completeness

3D reconstruction experiments are conducted for completeness evaluation. The number of reconstructed images and 3D points is set as criteria and statistic results are listed in Table 2. The results show that except for the MST, the other three methods could connect all images for the three datasets. By further checking the number of reconstructed points, some findings are identified. Firstly, point decrease ratios between the Full-TCN and the Reduced-TCN are less than 10%, 5% and 7% for dataset 1, dataset 2 and dataset 3, respectively. However, the corresponding decrease ratio of image pairs, represented in Figure 16(a), is more than 50%. It could be deduced that the pruned match pairs under SOC have much fewer matched features and have neglectable influence on the completeness of reconstruction. Secondly, for the MST-Expansion, point decrease ratios of the three test sites are about 39%, 36% and 33%, respectively. Although the number of reconstructed points is less than the Full-TCN and the Reduced-TCN, all images in the corresponding dataset are oriented and the reconstructed points could cover the whole test site as shown in Figure 18. Besides, much more dense point could can be generated through dense matching as presented in Figure A2.

Table 2. Number of reconstructed images and points for completeness comparison.

| Method | Site 1 | | Site 2 | | Site 3 | |
| --- | --- | --- | --- | --- | --- | --- |
| | #Images | #Points | #Images | #Points | #Images | #Points |
| Full-TCN | 320/320 | 152,075 | 390/390 | 243,667 | 750/750 | 248,128 |
| Reduced-TCN | 320/320 | 137,263 | 390/390 | 233,088 | 750/750 | 231,698 |
| MST | 128/320 | 38,581 | 75/390 | 34,714 | 48/750 | 4,819 |
| MST-Expansion | 320/320 | 91,622 | 390/390 | 155,561 | 750/750 | 167,465 |

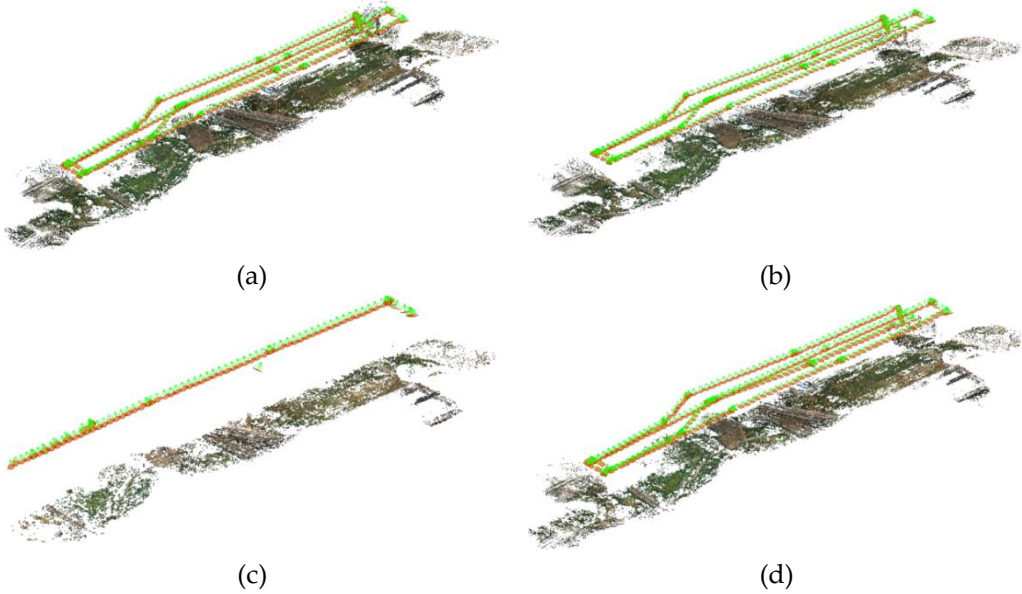

Figure 18. 3D scene reconstruction of dataset 1 guided by Full-TCN, Reduced-TCN, MST and MST-Expansion methods. (a) Full-TCN. (b) Reduced-TCN. (c) MST. (d) MST-Expansion. Image position and orientation have also been illustrated. The red rectangle stands for image plane and green line links projection center and one corner of image plane.

5.5.3. Accuracy

Bundle adjustment experiments without and with GCPs are executed for accuracy assessment. Without ground truth, bundle adjustment tests can be used to evaluate the relative accuracy of image connection network. Statistic results of the RMSE for the three methods are listed in Table 3. We can see that: 1) the RMSE of the MST-Expansion is smaller than that of other methods for each dataset. 2) For the MST-Expansion, the smallest RMSE is achieved in dataset 2 and the RMSE of dataset 1 is smaller than that of dataset 3.

The first finding could be explained by two reasons. On the one hand, less false matches are retained in the MST-Expansion because match pairs with smaller overlap area and larger intersection angle have been pruned for feature matching, which are the main sources of erroneous matches. On the other hand, the number of tracks involved in bundle adjustment is least than that of the other methods, which could further decrease the ratio of false matches. For the second finding, the main reason is that feature location accuracy is varying for images with different resolutions. Because sub-pixel location accuracy is achieved in the SIFT algorithm, features extracted from images with higher resolution have higher location accuracy, which inevitably affects the accuracy of bundle adjustment. By comparing RMSE with the corresponding GSD listed in Table 1, it is shown that the RMSE of bundle adjustment for each dataset is of nearly positive proportion to the image resolution.

Table 3. RMSE for bundle adjustment without GCPs (unit is pixel).

| Method | Site 1 | Site 2 | Site 3 |
| --- | --- | --- | --- |
| Full-TCN | 0.601 | 0.493 | 0.682 |
| Reduced-TCN | 0.583 | 0.469 | 0.664 |
| MST-Expansion | 0.447 | 0.337 | 0.541 |

With the aid of ground truth, geo-referencing accuracy assessment could be conducted. In this study, 43 GCPs evenly distributed in test site 2 are surveyed and prepared for geo-referencing accuracy analysis. Four GCPs, numbered as 7, 9, 35 and 37, are involved in absolute orientation, and all the others are utilized as check points (CPs) for geo-referencing accuracy comparison. After absolute orientation, the 3D model coordinate corresponding to each CP can be computed based on triangulation. Spatial distribution of orientation residuals, calculated by the subtraction between 3D model coordinates and CPs, is presented in Figure 19, in which spatial distribution of horizontal residuals of Full-TCN, Reduced-TCN and MST-Expansion is illustrated in Figure 19(a), Figure 19(c) and Figure 19(e), respectively. Similarly, spatial distribution of vertical residuals of the three methods is shown in Figure 19(b), Figure 19(d) and Figure 19(f), respectively. It is clearly shown that horizontal residual distribution is consistent for the three methods and no systematic errors could be obviously observed because of small residual magnitude and almost random residual direction. Although vertical residual distribution of the MST-Expansion is different from residual distribution of the Full-TCN and the Reduced-TCN, vertical residual magnitude is not greater than two times of the GSD value, where the GSD value is 3.67 cm for this dataset. Meanwhile, no systematic errors can be found from vertical residual distribution. The RMSEs in x, y and z directions for the three methods are shown in Figure 20. It is clear that competitive orientation accuracy could be achieved from the MST-Expansion method when compared with the orientation results of the Full-TCN and the Reduced-TCN.

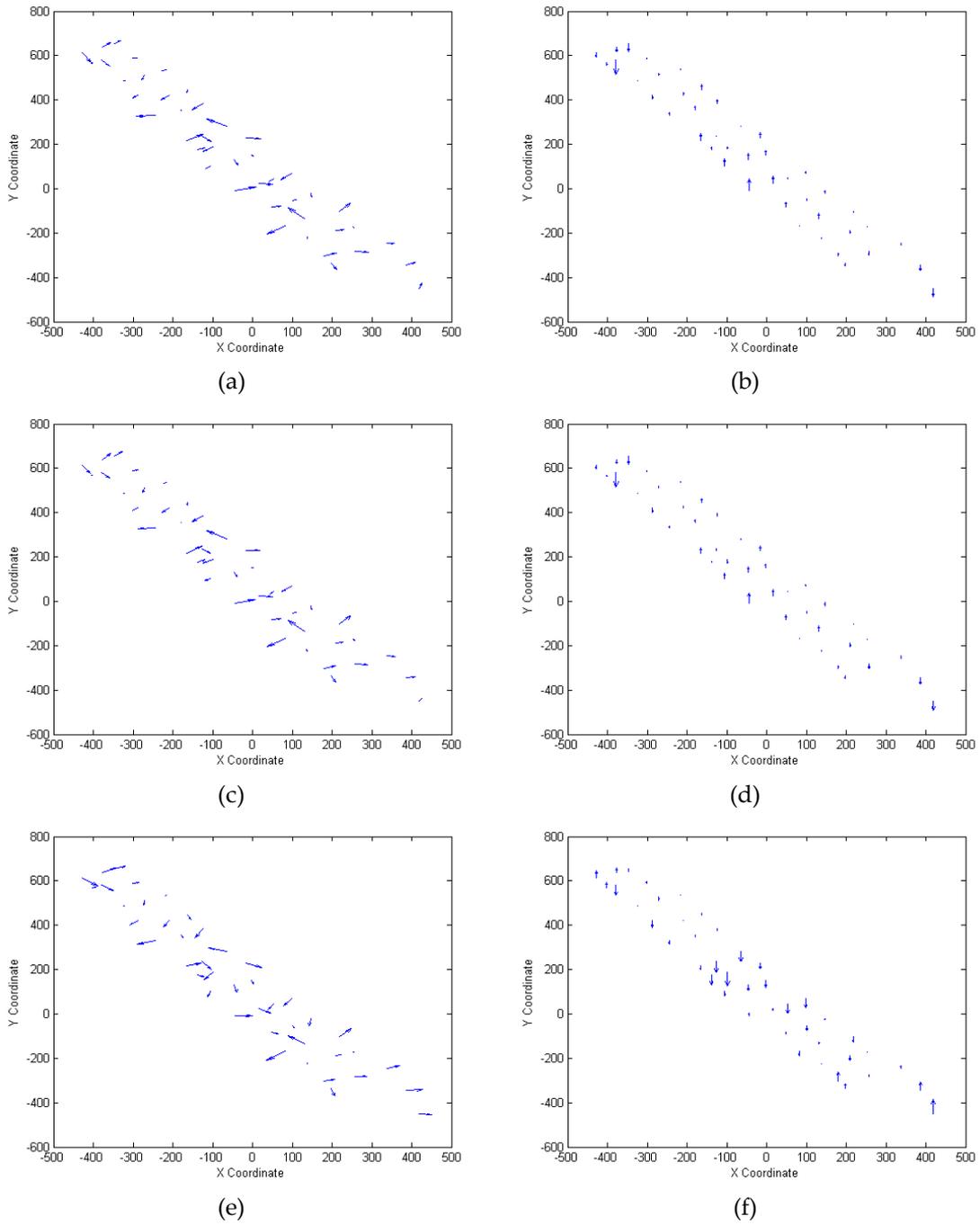

**Figure 19.** Residual spatial distribution for bundle adjustment of dataset 2 with GCPs. **(a)** and **(b)** horizontal and vertical residual distribution of the Full-TCN. **(c)** and **(d)** horizontal and vertical residual distribution of the Reduced-TCN. **(e)** and **(f)** horizontal and vertical residual distribution of the MST-Expansion.

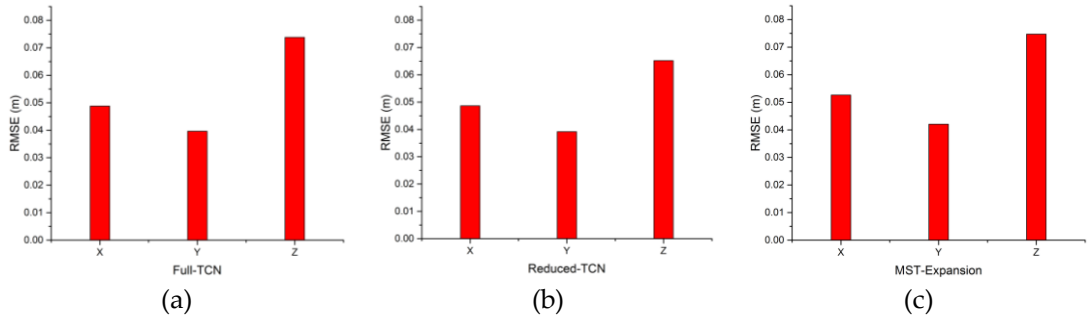

**Figure 20.** RMSE for bundle adjustment of dataset 2 with GCPs. **(a)** RMSE of the Full-TCN. **(b)** RMSE of the Reduced-TCN. **(c)** RMSE of the MST-Expansion.

*5.6. Compare MST-Expansion with MicMac*

MicMac is a free and open-source photogrammetric software package (Deseilligny and Clery, 2011). As one of the best software toolkits in photogrammetry, it provides a complete framework to compute and reconstruct 3D models based on principles of photogrammetry and compute vision, including image pair selection, SIFT feature extraction and matching, relative orientation and model geo-referencing. Among all functions of MicMac, the GrapheHom function is developed to search image pairs with overlap area by using image's prior position and orientation data, which aims for acceleration of feature matching. Then, the image pair file generated by the GrapheHom function is use as input of the Tapioca function, which makes use of multi-core parallel computation technology for feature extraction and matching. Finally, features extracted and matched by the Tapioca function are set as input to the orientation function Apero. With aid of GCPs, absolute orientation could be achieved for model geo-referencing. In this section, a comparison between MST-Expansion and MicMac is performed, including feature matching efficiency and bundle adjustment accuracy. All comparison experiments are conducted using dataset 2.

To compare feature matching efficiency, original images are firstly down-sampled to half size to accelerate feature extraction. Then, feature matching is executed for match pairs deduced from image's prior position and orientation, which is computed based on flight control data and camera mounting angle. The results of efficiency comparison for feature matching are listed in Table 4, in which less match pairs are retained in MST-Expansion and the matching time is no more than one three-hundredth of the time used in MicMac. Noticeably, time consumption ratio is dramatically different from ratio of image pair number. The main reason is that SiftGPU library with GPU acceleration is used in this study. However, the SIFT++ and the ANN are the default external libraries adopted in MicMac for feature extraction and matching, which does not utilize hardware acceleration. Thus, much higher efficiency can be achieved from MST-Expansion compared with MicMac.

To compare bundle adjustment accuracy, the same configuration for GCPs is utilized as described in section 5.5.3, where four GCPs, numbered as 7, 9, 35 and 37, are involved in absolute orientation. Accuracy comparison results are listed in Table 5, which shows that horizontal residuals, including mean value and RMSE, are almost equal. However, vertical residuals computed by MicMac are near two times of the value from the MST-Expansion. This could be explained by the insufficient camera self-calibration in MicMac, especially for focus length, because a subset of images in this dataset is used for the estimation of internal orientation parameters by using the calibration model termed RadialExtended.

Table 4. Efficiency comparison for feature matching between MST-Expansion and MicMac.

|  | Number of image pairs | Time used (in minites) |
|---|---|---|
| MicMac | 13,491 | 656.1 |
| MST-Expansion | 664 | 2.2 |

Table 5. Accuracy comparison for bundle adjustment between MST-Expansion and MicMac.

|  | Mean | | | RMSE | | |
|---|---|---|---|---|---|---|
|  | \|X\|(m) | \|Y\|(m) | \|Z\|(m) | X(m) | Y(m) | Z(m) |
| MicMac | 0.046 | 0.041 | 0.123 | 0.056 | 0.051 | 0.150 |
| MST-Expansion | 0.045 | 0.038 | 0.062 | 0.053 | 0.042 | 0.075 |

## 6. Discussion

This research proposes the SRC-InterTest algorithm for match pair selection and the MST-Expansion algorithm for match graph extraction to achieve an efficient SfM solution for oblique UAV images. The experimental results confirm that the computational costs are dramatically decreased with the aid of the proposed algorithms. In the literature, some other strategies have also been reported for an efficient SfM, including time sequential constrain (AliAkbarpour et al., 2015), vocabulary tree based similarity score (Irschara et al., 2011) and fixed distance threshold (Rupnik et al., 2013), etc. Compared with these methods, the proposed solution has the following advantages.

First, the SRC-InterTest algorithm is achieved by the further exploration of SDC and SOC constrains. SDC enables match pair selection with linear time complexity and SOC decreases the risk of incorrect matches involved in image orientation, which can be deduced from the experimental results in section 5.2 and 5.5.3. The proposed match pair selection strategy is more self-adaptive and more efficient because it is independent of a fixed distance threshold for neighbor searching and can also avoid exhausted intersection tests. Second, both overlap area and intersection angle are used for the computation of edge weight within the TCN construction, which encodes the influence of oblique angles to quantify the importance of one image pair. As verified in the experiment of section 5.3, the involvement of oblique angle ensures a complete reconstruction. Hence, the MST-Expansion algorithm is adapted to both vertical and oblique datasets. In addition, MST-Expansion is a non-iterative and two-stage algorithm based on local structure analysis for TCN expansion. Thus, the computational costs for match graph extraction become negligible even for large datasets. Finally, the proposed SfM method does not rely on other data sources except for the GNSS/IMU data from flight control system and camera mounting angles of an oblique imaging system, which expands considerably the range of its application in the field of UAV photogrammetry.

Through the analysis of efficiency presented in section 5.5.1, it is clearly show that much less match pairs are retained from the MST-Expansion algorithm, which accelerates feature matching with the speedup ratios ranging from 19 to 35 for the three datasets. In addition, with the increase of image overlap degree, higher efficiency could be achieved because much more redundant match pairs would be existed. Meanwhile, the RMSEs estimated from both relative and absolute bundle adjustments confirm that, compared with the other methods, the proposed SfM solution can achieve competitive orientation accuracy even with much fewer match pairs involved in feature matching. Although images in each datasets are successfully connected, the number of reconstructed 3D points from the MST-expansion algorithm is less than the one from the other methods. Therefore, further improvements can aim to resume much more3D point through the global consistency analysis (Shen et al., 2016).

As verified by the comparison in section 5.6, efficiency has been dramatically improved by using the SiftGPU algorithm. However, two problems relating to feature extraction and matching need to be considered. First, the default configuration of the SiftGPU algorithm is used in this study, where the maximum image size is limited to 3200 pixels. With image resolution increasing, the SiftGPU algorithm cannot directly provide a solution to feature extraction and matching for large images because of either a loss of detail or the reduction of feature number. To cope with this situation, one possible way is to split large images into tiles and perform feature extraction and matching on each tile (Sun et al., 2014; Xu et al., 2016). Second, feature matching between oblique images is extraordinary difficult because of different appearances mainly caused by perspective deformation, which can be observed from the completeness analysis when TCN weighted without oblique angles. To deal with this problem, the GNSS/IMU data can also be used to horizontally rectify original images to relieve the perspective deformations (Hu et al., 2015).

## 7. Conclusions

In this paper, we present the SRC-InterTest algorithm and the MST-Expansion algorithm to achieve an efficient SfM solution for oblique UAV images. The SRC-InterTest algorithm is used for match pair selection without the dependency upon a fixed threshold for neighbor searching and without exhausted intersection test. The MST-Expansion algorithm is proposed for match graph extraction from initial match pairs, which improves the efficiency of image matching for oblique UAV images with high resolution and overlap degree. Finally, from the aspects of efficiency, completeness and accuracy, the proposed SfM solution is evaluated by intensive comparison and analysis using three oblique UAV datasets captured with different oblique multi-camera systems. As shown in the experiments, the SRC-InterTest algorithm can efficiently select image pairs with linear time complexity; the MST-Expansion algorithm can dramatically reduce computational costs of image matching and achieve an efficient SfM solution. Meanwhile, competitive orientation accuracy from BA tests is obtained when compared with other methods. For orientation of oblique UAV images, the proposed solution in this paper can be a more efficient method.

**Acknowledgment**

The authors would like to thank authors who have made their algorithms of SiftGPU, CMVS/PMVS, MicMac as free and open-source software packages, which is really helpful to the research in this paper.

**Appendix A**

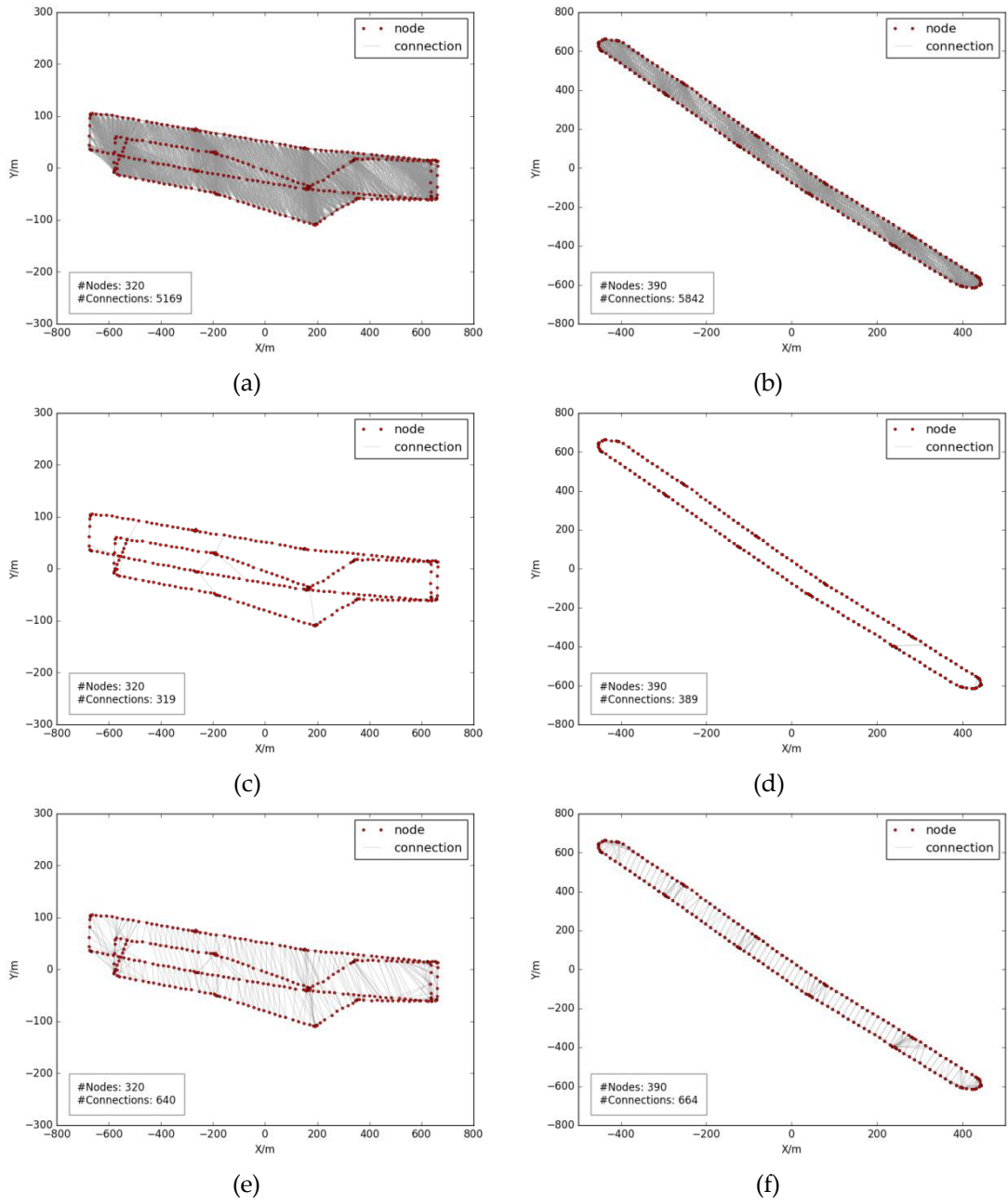

**Figure A1.** Match graphs for dataset 1 and dataset 2. **(a)** and **(b)** match graphs generated from Reduced-TCN; **(c)** and **(d)** match graphs generated from MST; **(e)** and **(f)** match graphs generated from MST-Expansion.

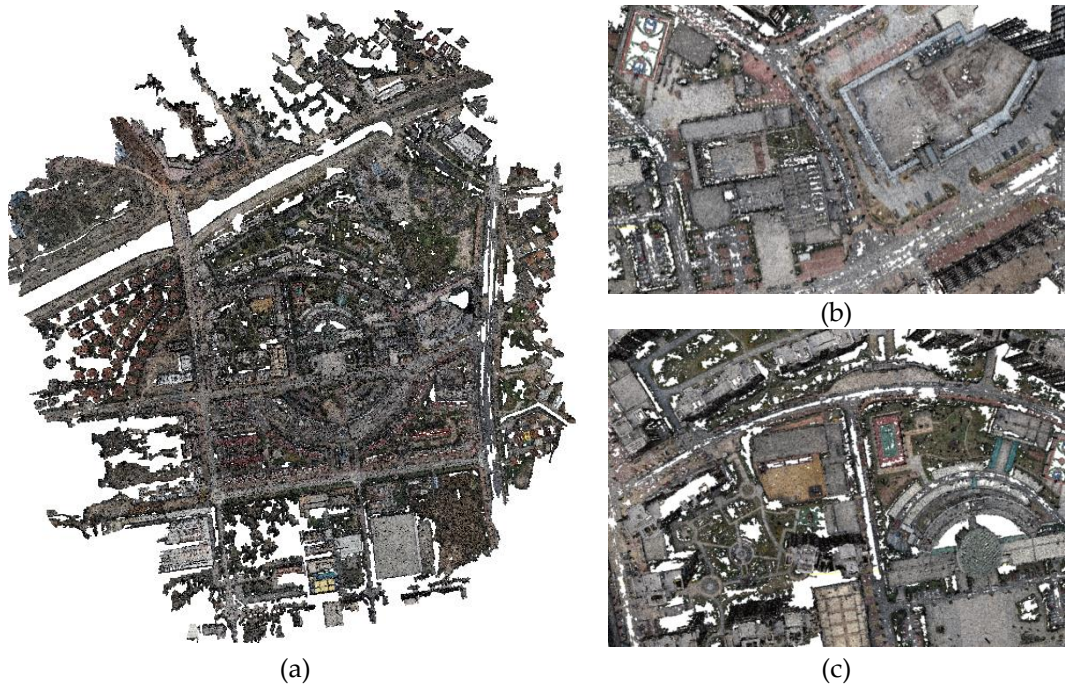

(a)　　　　　　　　　　　　　　　　(c)

**Figure A2.** Dense matching point cloud of dataset 3 based on image orientation results generated by the proposed SfM (structure from motion) method: **(a)** overall scene, **(b)** and **(c)** local detailed scenes. This point cloud is generated using the open-source software package PMVS (Furukawa and Ponce, 2012).

**References**

1. Agarwal, S., Snavely, N., Seitz, S.M., Szeliski, R., 2010. Bundle adjustment in the large, Computer Vision–ECCV 2010. Springer, pp. 29-42.
2. Aicardi, I., Chiabrando, F., Grasso, N., Lingua, A.M., Noardo, F., Spanò, A., 2016a. UAV photogrammetry with oblique images: First analysis on data acquisition and processing. International Archives of the Photogrammetry, Remote Sensing and Spatial Information Sciences 41, 835-842.
3. Aicardi, I., Nex, F., Gerke, M., Lingua, A.M., 2016b. An Image-Based Approach for the Co-Registration of Multi-Temporal UAV Image Datasets. Remote Sensing 8, 779.
4. AliAkbarpour, H., Palaniappan, K., Seetharaman, G., 2015. Fast Structure from Motion for Sequential and Wide Area Motion Imagery, 2015 IEEE International Conference on Computer Vision Workshop (ICCVW), pp. 1086-1093.
5. Alsadik, B., Gerke, M., Vosselman, G., Daham, A., Jasim, L., 2014. Minimal camera networks for 3D image based modeling of cultural heritage objects. Sensors-Basel 14, 5785-5804.
6. Andrew, A.M., 1979. Another efficient algorithm for convex hulls in two dimensions. Information Processing Letters 9, 216-219.
7. Chidburee, P., Mills, J.P., Miller, P.E., Fieber, K.D., 2016. Towards a Low-Cost Real-Time Photogrammetric Landslide Monitoring System Utilising Mobile and Cloud Computing Technology. Int. Arch. Photogramm. Remote Sens. Spatial Inf. Sci. XLI-B5, 791-797.
8. Colomina, I., Molina, P., 2014. Unmanned aerial systems for photogrammetry and remote sensing: A review. ISPRS Journal of Photogrammetry and Remote Sensing 92, 79-97.
9. Cover, T., Hart, P., 1967. Nearest neighbor pattern classification. IEEE Transactions on Information Theory 13, 21-27.
10. Deseilligny, M.P., Clery, I., 2011. Apero, an open source bundle adjustment software for automatic calibration and orientation of set of images. ISPRS-International Archives of the Photogrammetry, Remote Sensing and Spatial Information Sciences 38, 5.

11. Fischler, M.A., Bolles, R.C., 1981. Random sample consensus: a paradigm for model fitting with applications to image analysis and automated cartography. Communications of the ACM 24, 381-395.
12. Furukawa, Y., Ponce, J., 2012. Patch-based Multi-view Stereo Software (PMVSVersion 2) < http://francemapping.free.fr/Portfolio/Prog3D/PMVS2.html > (Accessed 2017 April 6).
13. Gerke, M., 2009. Dense matching in high resolution oblique airborne images. Int. Arch. Photogramm. Remote Sens. Spat. Inf. Sci 38, W4.
14. Gonçalves, J.A., Henriques, R., 2015. UAV photogrammetry for topographic monitoring of coastal areas. ISPRS Journal of Photogrammetry and Remote Sensing 104, 101-111.
15. Hartmann, W., Havlena, M., Schindler, K., 2015. Recent developments in large-scale tie-point matching. ISPRS Journal of Photogrammetry and Remote Sensing.
16. Havlena, M., Torii, A., Pajdla, T., 2010. Efficient structure from motion by graph optimization, European Conference on Computer Vision. Springer, pp. 100-113.
17. Heinly, J., Schonberger, J.L., Dunn, E., Frahm, J.-M., 2015. Reconstructing the world* in six days*(as captured by the yahoo 100 million image dataset), Proceedings of the IEEE Conference on Computer Vision and Pattern Recognition, pp. 3287-3295.
18. Hu, H., Zhu, Q., Du, Z., Zhang, Y., Ding, Y., 2015. Reliable Spatial Relationship Constrained Feature Point Matching of Oblique Aerial Images. Photogrammetric Engineering & Remote Sensing 81, 49-58.
19. Ippoliti, E., Meschini, A., Sicuranza, F., 2015. Structure from Motion Systems for Architectural Heritage. a Survey of the Internal Loggia Courtyard of Palazzo Dei Capitani, Ascoli Piceno, Italy. Int. Arch. Photogramm. Remote Sens. Spatial Inf. Sci. XL-5/W4, 53-60.
20. Irschara, A., Hoppe, C., Bischof, H., Kluckner, S., 2011. Efficient structure from motion with weak position and orientation priors, CVPR 2011 WORKSHOPS, pp. 21-28.
21. Jiang, S., Jiang, W., Huang, W., Yang, L., 2017. UAV-Based Oblique Photogrammetry for Outdoor Data Acquisition and Offsite Visual Inspection of Transmission Line. Remote Sensing 9, 278.
22. Kruskal, J.B., 1956. On the shortest spanning subtree of a graph and the traveling salesman problem. Proceedings of the American Mathematical society 7, 48-50.
23. Lin, Y., Jiang, M., Yao, Y., Zhang, L., Lin, J., 2015. Use of UAV oblique imaging for the detection of individual trees in residential environments. Urban Forestry & Urban Greening 14, 404-412.
24. Lowe, D.G., 2004. Distinctive image features from scale-invariant keypoints. International journal of computer vision 60, 91-110.
25. Matsuoka, R., Nagusa, I., Yasuhara, H., Mori, M., Katayama, T., Yachi, N., Hasui, A., Katakuse, M., Atagi, T., 2012. Measurement of Large-Scale Solar Power Plant by Using Images Acquired by Non-Metric Digital Camera on Board Uav. ISPRS - International Archives of the Photogrammetry, Remote Sensing and Spatial Information Sciences 39, 435-440.
26. Mizotin, M., Krivovyaz, G., Velizhev, A., Chernyavskiy, A., Sechin, A., 2010. Robust matching of aerial images with low overlap. International Archives of the Photogrammetry, Remote Sensing and Spatial Information Sciences 38, 13-18.
27. Nex, F., Remondino, F., 2014. UAV for 3D mapping applications: a review. Applied Geomatics 6, 1-15.
28. Remondino, F., Gerke, M., 2015. Oblique Aerial Imagery–A Review. Proc. Week, 75-83.
29. Rupnik, E., Nex, F., Remondino, F., 2013. Automatic orientation of large blocks of oblique images. Int. Archives of Photogrammetry, Remote Sensing and Spatial Information Sciences 40, 1.
30. Rupnik, E., Nex, F., Remondino, F., 2014. Oblique multi-camera systems-orientation and dense matching issues. The International Archives of Photogrammetry, Remote Sensing and Spatial Information Sciences 40, 107.
31. Sameer, A., Keir, M., 2010. Ceres Solver <http://ceres-solver.org > (Accessed 2017 April 6).
32. Shen, T., Zhu, S., Fang, T., Zhang, R., Quan, L., 2016. Graph-based consistent matching for structure-from-motion, European Conference on Computer Vision. Springer, pp. 139-155.
33. Snavely, N., Seitz, S.M., Szeliski, R., 2006. Photo tourism: exploring photo collections in 3D, ACM transactions on graphics (TOG). ACM, pp. 835-846.
34. Snavely, N., Seitz, S.M., Szeliski, R., 2008. Skeletal graphs for efficient structure from motion, 2008 IEEE Conference on Computer Vision and Pattern Recognition, pp. 1-8.

35. Sun, Y., Sun, H., Yan, L., Fan, S., Chen, R., 2016. RBA: Reduced Bundle Adjustment for oblique aerial photogrammetry. ISPRS Journal of Photogrammetry and Remote Sensing 121, 128-142.

36. Sun, Y., Zhao, L., Huang, S., Yan, L., Dissanayake, G., 2014. L2-SIFT: SIFT feature extraction and matching for large images in large-scale aerial photogrammetry. ISPRS Journal of Photogrammetry and Remote Sensing 91, 1-16.

37. Vetrivel, A., Gerke, M., Kerle, N., Vosselman, G., 2015. Identification of damage in buildings based on gaps in 3D point clouds from very high resolution oblique airborne images. ISPRS journal of photogrammetry and remote sensing 105, 61-78.

38. Westoby, M., Brasington, J., Glasser, N., Hambrey, M., Reynolds, J., 2012. 'Structure-from-Motion' photogrammetry: A low-cost, effective tool for geoscience applications. Geomorphology 179, 300-314.

39. Wu, C., 2007. SiftGPU: A GPU implementation of scale invariant feature transform (SIFT).

40. Xu, Z., Wu, L., Gerke, M., Wang, R., Yang, H., 2016. Skeletal camera network embedded structure-from-motion for 3D scene reconstruction from UAV images. ISPRS Journal of Photogrammetry and Remote Sensing 121, 113-127.

41. Zhang, R., Schneider, D., Strauß, B., 2016. Generation and Comparison of TLS and SfM based 3D Models of Solid Shapes in Hydromechanical Research. Int. Arch. Photogramm. Remote Sens. Spatial Inf. Sci. XLI-B5, 925-929.

42. Zhu, Q., Jiang, W., Zhang, J., 2015. Feature line based building detection and reconstruction from oblique airborne imagery. The International Archives of Photogrammetry, Remote Sensing and Spatial Information Sciences 40, 199.